\documentclass[twoside,11pt]{article}
\usepackage{jmlr2e}
\usepackage{hyperref}       
\usepackage{url}            
\usepackage{booktabs}       
\usepackage{amsfonts}       
\usepackage{nicefrac}       
\usepackage{microtype}      
\usepackage{amsmath}
\usepackage{graphicx}
\usepackage{xcolor}
\usepackage{epstopdf}
\usepackage{wrapfig}
\usepackage{subcaption}

\graphicspath{ {images/} }

\newcommand{\R}{\mathbb{R}}

\newcommand{\norm}[1]{\lVert{#1}\rVert}

\newcommand{\bmat}[1]{\begin{bmatrix}#1\end{bmatrix}}

\newtheorem{thm}{Theorem}
\newtheorem{prop}[thm]{Proposition}
\newtheorem{lem}[thm]{Lemma}
\newtheorem{defn}[thm]{Definition}
\newtheorem{corr}[thm]{Corollary}
\newcommand{\ip}[2]{\langle #1, #2 \rangle}
\newcommand{\mcl}[1]{\mathcal{#1}}
\newcommand{\half}{\frac{1}{2}}

\newcommand{\N}{\mathbb{N}}

\firstpageno{1}

\usepackage{lastpage}
\jmlrheading{20}{2019}{1-\pageref{LastPage}}{7/19}{10/19}{19-594}{Brendon K. Colbert and Matthew M. Peet}
\ShortHeadings{A NEW CLASS OF UNIVERSAL KERNEL FUNCTIONS}{Colbert and Peet}

\begin{document}

\title{A Convex Parametrization of a New Class of Universal Kernel Functions}

\author{\name Brendon K. Colbert \email brendon.colbert@asu.edu \\
       \addr Department of Mechanical and Aerospace Engineering\\
       Arizona State University\\
       Tempe, AZ 85281-4322, USA
       \AND
       \name Matthew M.\ Peet \email mpeet@asu.edu \\
       \addr Department of Mechanical and Aerospace Engineering\\
       Arizona State University\\
       Tempe, AZ 85281-1776, USA}

\editor{Mehryar Mohri}

\maketitle

\begin{abstract}%
The accuracy and complexity of kernel learning algorithms is determined by the set of kernels over which it is able to optimize. An ideal set of kernels should: admit a linear parameterization (tractability); be dense in the set of all kernels (accuracy); and every member should be universal so that the hypothesis space is infinite-dimensional (scalability). Currently, there is no class of kernel that meets all three criteria - e.g. Gaussians are not tractable or accurate; polynomials are not scalable. We propose a new class that meet all three criteria - the Tessellated Kernel (TK) class. Specifically, the TK class: admits a linear parameterization using positive matrices; is dense in all kernels; and every element in the class is universal. This implies that the use of TK kernels for learning the kernel can obviate the need for selecting candidate kernels in algorithms such as SimpleMKL and parameters such as the bandwidth.  Numerical testing on soft margin Support Vector Machine (SVM) problems show that algorithms using TK kernels outperform other kernel learning algorithms and neural networks. Furthermore, our results show that when the ratio of the number of training data to features is high, the improvement of TK over MKL increases significantly.
\end{abstract}


\section{Introduction}
This paper addresses the problem of the automated selection of an optimal kernel function for a given kernel-based machine learning problem (e.g. soft margin SVM). Kernel functions implicitly define a linear parametrization of nonlinear candidate maps $y=f(x)$ from vectors $x$ to scalars $y$. Specifically, for a given kernel, the `kernel trick' allows optimization over a set of candidate functions in the kernel-associated hypothesis space without explicit representation of the space itself. The kernel selection process, then, is critical for determining the class of hypothesis functions and, as a result, is a well-studied topic with common kernels including polynomials, Gaussians, and many variations of the Radial Basis Function. In addition, specialized kernels include string kernels as in~\cite{lodhi_2002,eskin_2003}, graph kernels as in~\cite{gartner_2003}, and convolution kernels~as in~\cite{haussler_1999,collins_2002}. The kernel selection process heavily influences the accuracy of the resulting fit and hence significant research has gone into the optimization of these kernel functions in order to select the hypothesis space which most accurately represents the underlying physical process.

Recently, there have been a number of proposed kernel learning algorithms. For support vector machines, the methods proposed in this paper are heavily influenced by the SDP approach proposed by~\cite{lanckriet_2004} which directly imposed kernel matrix positivity on a subspace defined by the linear combination of candidate kernel functions. There have been several extensions of the SDP approach, including the hyperkernel method of~\cite{ong2005learning}. However, because of the complexity of semidefinite programming, more recent work has focused on alignment methods for MKL as in, e.g.~\cite{cortes_2012} or gradient methods for convex and non-convex parameterizations of positive linear combinations of candidate kernels, such as SimpleMKL in ~\cite{rakotomamonjy_2008} or the several variations in~\cite{SHOGUN}. These MKL methods rely on kernel operations (addition, multiplication, convolution) to generate large numbers of parameterized kernel functions as in~\cite{cortes2009learning}. Examples of non-convex parameterizations include GMKL as introduced in~\cite{jain_2012}, and LMKL as introduced in~\cite{gonen_2008}.  Work focused on regularization includes the group sparsity metric defined in~\cite{subrahmanya_2010} and the enclosing ball approach in~\cite{gai2010learning}. See, e.g.~\cite{gonen2011multiple} for a comprehensive review of MKL algorithms.

In this paper, we focus on the class of ``Universal Kernels'' formalized in~\cite{micchelli_2006}. For a given compact metric space (input space), $\mcl X$, it is said that a function $k:\mcl X \times \mcl X \rightarrow \R$ is a Positive Kernel (PK) if for any $N \in \N$ and any $\{x_i\}_{i=1}^N \subset \mcl X$, the matrix defined elementwise by $K_{ij}=k(x_i,x_j)$ is symmetric and Positive SemiDefinite (PSD).

\begin{defn}
A kernel $k:\mcl X \times \mcl X \rightarrow \R$ is said to be \textit{universal} on the compact metric space $\mcl X$ if it is continuous and there exists an inner-product space $\mcl W$ and feature map, $\Phi : \mcl X \rightarrow \mcl W$ such that $k(x,y)=\ip{\Phi(x)}{\Phi(y)}_{\mcl W}$ and where the unique Reproducing Kernel Hilbert Space (RKHS),
\[
\mcl H:=\{f\;:\;f(x)=\ip{v}{\Phi(x)},\; v \in \mcl W\}
\]
with associated norm $\norm{f}_{\mcl H} :=\inf_{v}\{\norm{v}_{\mcl W}\;:\; f(x)=\ip{v}{\Phi(x)}\}$ is dense in $\mcl C(\mcl X):=\{f \,: \, \mcl X \rightarrow \R \;:\; $f$\, \text{ is continuous}\}$ where $\norm{f}_{\mcl C}:=\sup_{x\in X} |f(x)|$.
\end{defn}
Note that for any given PD kernel, $\mcl H$ exists, is unique, and can be characterized (as described in~\cite{sun_2005}) using the Riesz representation theorem as the closure of $\text{span}\{k(y,\cdot)\,:\,y \in \mcl X \}$ with inner product defined for any $f(x)=\sum_{i=1}^n c_i k(y_i,x)$ and $g(x)=\sum_{i=1}^m d_i k(z_i,x)$ as
\[
\ip{f}{g}_{\mcl H}:=\sum_{i=1}^n \sum_{j=1}^m c_i d_j k(y_i,z_j).
\]
%
%
%
%
Universal kernels are preferred when large amounts of data are available, due to the fact that the dimension of the hypothesis space increases for every additional data point - resulting in the ability to construct highly specialized and accurate classifiers.
The most well-known example of a universal kernel is the Gaussian (generalized in~\cite{zanaty_2011}).  However, many other common kernels are not universal, including, significantly, the polynomial class of kernels. This is significant because generalized polynomial kernels (Eq.~\eqref{polyKernel}) are dense in all kernels and admit a linear parameterization using a monomial basis, while Gaussian kernels (which are universal, but not dense in all kernels) do not.

\paragraph{The Class of Tessellated Kernels (TK)} In this paper we propose a new class of kernel functions (called Tessellated Kernels) which are not polynomials, yet which are defined by polynomials and admit a linear parametrization. These kernels define classifiers on a tessellated domain, each sub-domain (or tile) of which is a hyper-rectangle with vertices defined by the \textit{input data} - $\{ x_i \}_{i=1}^m$. In this way, each data point further divides any tiles within which any of its features lie, resulting in increasing numbers of disjoint tiles. The classifier itself, then, is piecewise polynomial - being polynomial when restricted to any particular tile.

TK kernels have three important properties which make them uniquely well-suited for kernel learning problems. First, these kernels admit a linear parameterization using positive semidefinite matrices - meaning we can use convex optimization to search over the entire class of such kernels (tractability), which is proven in Corollary~\ref{corr} and implemented in Optimization Problem~\eqref{SDPOPT}.  This is like the class of generalized polynomial kernels (See Eq.~\eqref{polyKernel}) yet unlike other universal kernel classes such as the Gaussian/RBF, wherein the bandwidth parameter appears in the exponential.  Second, the TK class is dense in all kernels (accuracy), meaning there exists a TK kernel that can approximate any given kernel arbitrarily well. This is like the generalized polynomial class yet unlike the Gaussian/RBF class, wherein the resulting kernel matrix is restricted to having all positive elements. Third, any kernel of the TK class has the universal property (scalability).  This is like the Gaussian/RBF class and unlike the generalized polynomial kernels, none of which are universal. The TK class is thus unique in that no other currently known class of kernel functions has all three properties of tractability, accuracy, and scalability. Finally, we demonstrated through extensive numerical testing that kernel learning using TK kernels significantly outperforms other kernel learning algorithms in terms of accuracy.
%
%

The paper is organized as follows. In Section \ref{sec:2} we provide an overview of the MKL problem.  Section \ref{sec:3} proposes a framework by which the class of TK kernels can be parameterized by positive matrices. Section \ref{sec:4} proves general properties such as universality for every member of the class of TK kernels.  Sections \ref{sec:SDP} and \ref{sec:6} show how the class of TK kernels can be rigorously incorporated into the SDP MKL framework and into SimpleMKL's framework respectively.  In Section \ref{sec:7} we discuss the complexity of incorporating TK kernels into both the SDP MKL framework and the SimpleMKL framework.  Finally in Section \ref{sec:8} we provide numerical results that illustrate improved performance using TK kernels on a number of UCI repository data sets.


\section{Formulation of the Kernel Learning Problem}\label{sec:2}\vspace{-3mm}
We begin this section by posing the kernel-learning problem as a convex optimization problem for the particular case of the 1-norm soft margin support vector machine. Next, in Subsections A and B, we present two standard algorithms for solving the kernel learning problem. These algorithms are general in the sense that they apply to any given linear parameterization of kernel functions. The adaptation of these algorithms to the special case of TK kernels will then be described in Section~\ref{sec:3}.

Suppose we are given a set of $m$ \textit{training data} points $\{x_i\}^m_{i=1} \subset \R^{n}$, each with associated \textit{label} $y_i \in \{-1,1\}$ for $i =1, \cdots, m$. For a given ``penalty'' parameter $C \in \R^+$, we define the primal version of the \textit{linear} 1-norm soft margin problem as \vspace{-1mm}
\begin{align}
\min_{w \in \R^n,~\zeta \in \R^{m},~b \in \R} & \quad \frac{1}{2} w^Tw + C \sum_{i=1}^m \zeta_i \label{eqn:SVMlinear} \\
\text{s.t.} & \quad y_i( w^Tx_i+ b) \geq 1 - \zeta_i,\quad \zeta_i\ge 0, \notag 
\end{align}
where the learned map (\textit{classifier}) from inputs to outputs is then $f: \R^n \rightarrow \{-1,1\}$ where
\[ f(x) = \text{sign}(w^Tx + b). \]
If we desire the classifier to be a nonlinear function, we may introduce a positive kernel function, $k$.
\begin{defn}
We say a function $k:  Y \times Y \rightarrow \R$ is a {\bf positive kernel function} if
\[
\int_{Y} \int_{Y} f(x)k(x,y)f(y)dxdy\ge 0
\]
for any function $f \in L_2[Y]$.
\end{defn}
For any given positive kernel $k$ we may associate a $\Phi$ such that $k(x,y) = \ip{\Phi(x)}{\Phi(y)}$. In this case Optimization Problem~\eqref{eqn:SVMlinear} might be posed as
\begin{align}
\min_{w \in \R^n,~\zeta \in \R^{m},~b \in \R} & \quad \frac{1}{2} w^T w + C \sum_{i=1}^m \zeta_i \label{eqn:SVM} \\
\text{s.t.} & \quad y_i(\ip {w} {\Phi(x_i)}+b) \geq 1 - \zeta_i,\quad \zeta_i\ge 0.  \notag
\end{align}
Given a solution, the classifier would be
\[
f(z) = \text{sign}\left( \ip{w}{\Phi(z)} + b\right).
\]
Although the primal form of SVM has certain advantages - see~\cite{rahimi_2008}, it is ill-suited to kernel learning. For this reason, we consider the dual formulation,
\begin{align}  \label{KernelSVM}
\max_{\alpha \in \R^m} & \quad \sum_{i=1}^m \alpha_i - \frac{1}{2} \sum_{i=1}^m\sum_{j=1}^m \alpha_i \alpha_j y_i y_j \ip{\Phi(x_i)}{\Phi(x_j)} \\ \nonumber
\text{s.t.} & \quad \sum_{i=1}^m \alpha_iy_i = 0, \quad 0 \leq \alpha_i \leq C ~~ \forall ~~ i = 1,...,m.
\end{align}
In this case we may eliminate $\Phi$ from the optimization problem using $\ip{\Phi(x_i)}{\Phi(x_j)} = k(x_i,x_j)$ where the elements $k(x_i,x_j)$ define the kernel matrix. In this case, the resulting classifier is only a function of $k$ and becomes
\[
 f(z) = \text{sign}\left(\sum_{i=1}^m \alpha_iy_ik(x_i,z) + b\right).
 \]
Note that $b$ can be found a posteriori as the average of $y_j- \sum_{i=1}^m \alpha_iy_ik(x_j,x_i)$ for all $j$ such that $0< \alpha_j < C$ - See \cite{LearningWithKernels}.  This implies that the primal variable $w$ is not explicitly required for the calculation of $b$, and that the resulting learned  classifier, $f$, may be expressed solely in terms of $\alpha$ and the kernel function.

Commonly used positive kernel functions include the gaussian kernel $k_1(x,y) = \text{e}^{(-\beta||x-y||^2)}$, where $\beta$ is the bandwidth (and must be chosen a priori) and the polynomial kernel $k_2(x,y)=(1+x^Ty)^d$ where $d$ is the degree of the polynomial.

Unfortunately Optimization Problem \ref{KernelSVM} requires that the kernel function, $k(x,y)$, be chosen a priori, a choice which significantly influences the accuracy of the resulting classifier $f$.  We therefore alter the optimization problem by considering the kernel itself to be an optimization variable, constrained to lie in a given convex set of candidate positive kernel functions, $\mcl K$. In this case, we have the following convex optimization problem.
\begin{align} \label{OptimalKernelSVM1}
\min_{k \in \mcl K} \max_{~\alpha \in \R^m} & \quad \sum_{i=1}^m \alpha_i - \frac{1}{2} \sum_{i=1}^m\sum_{j=1}^m \alpha_i \alpha_j y_i y_j k(x_i,x_j) \\ \nonumber
\text{s.t.} & \quad \sum_{i=1}^m \alpha_iy_i = 0, \quad 0 \leq \alpha_i \leq C ~~ \forall ~~ i = 1,...,m
\end{align}
Having formulated the kernel learning problem, we now present two standard approaches to parameterizing the set of candidate kernels, $\mcl K$, and solving the resulting convex optimization problem.

\subsection{SDP-Based Kernel Learning Using Positive Kernel Matrices}
We first consider the method of~\cite{lanckriet_2004}, wherein positive matrices were used to parameterize $\mcl K$ for a given set of candidate kernels $\{k_i\}_{i=1}^l$ as
\[
\mcl K:= \left\{ k(x,y)=\sum_{i=1}^l \mu_i k_i(x,y) :\; \mu \in \R^l,\; K_{ij}=k(x_i,x_j),\; K\geq 0\right\},
\]
where the $\{x_i\}^m_{i=1} \subset \R^{n}$ are the training points of the SVM problem and the $k_i$ were chosen a priori to be, for instance, Gaussian and polynomial kernels. It is significant to note that the PSD constraint on the kernel matrix $K$, enforces that the kernel matrix is PSD for the set of training data, but does not necessarily enforce that the kernel function itself is PD - meaning that kernels in $\mcl K$ are not necessarily positive kernels.

Using this parameterized $\mcl K$, the kernel optimization problem for the 1-norm soft margin support vector machine was formulated in~\cite{lanckriet_2004} as the following semi-definite program, where $e$ is the vector of all ones.
\begin{align} \label{SDP}
&\underset{\mu \in \R^l,~t \in \R,~\gamma \in \R,~\nu \in \R^{m},~\delta \in \R^m}{\text{min}} \hspace{20mm} t \hspace{10mm} \\ \nonumber
 & \text{subject to:}  \quad ~ \begin{pmatrix}
  G &  e + \nu - \delta + \gamma y \\ \nonumber
  (e + \nu - \delta + \gamma y)^T & t-\frac{2}{m \lambda } \delta^T  e
 \end{pmatrix} \geq 0\\ \nonumber
&  \hspace{10mm}  \nu \geq 0,\qquad  \delta \geq 0,\qquad  G_{ij}=k(x_i,x_j) y_i y_j\\
&  \hspace{10mm}k(x,y)=\sum_{i=1}^l \mu_i k_i(x,y) \notag
\end{align}
Note that here the original constraint $K\ge 0$ in $\mcl K$ has been replaced by an equivalent constraint on $G$. This problem can now be solved using well-developed interior-point methods as in~\cite{alizadeh_1998} with implementations such as MOSEK in~\cite{mosek}.

In Optimization Problem \eqref{SDP}, the size of the SDP constraint is $(m+1) \times (m+1)$ which is problematic in that the complexity of the resulting SDP grows as a polynomial in the number of training data.  Methods that do not require this large semi-definite matrix constraint are explored next. 

\subsection{Kernel Learning Using MKL}
In this subsection, we again take a set of basis kernels $\{k_i\}_{i=1}^l$ and consider the set of positive linear combinations,
\begin{align}\label{simpleMKLSet}
\mcl K:= \left\{k\;:\; k(x,y)=\sum_{i=1}^l \mu_i k_i(x,y), \; \mu_i \ge 0\right\}.
\end{align}
Any element of this set is a positive kernel, replacing the matrix positivity constraint by a LP constraint.
\begin{align}
\min_{\mu \ge 0} \max_{~\alpha \in \R^m} & \quad \sum_{i=1}^m \alpha_i - \frac{1}{2} \sum_{i=1}^m\sum_{j=1}^m\sum_{k=1}^l \mu_k \alpha_i \alpha_j y_i y_j k_k(x_i,x_j) \notag \\ \nonumber
\text{s.t.} & \quad \sum_{i=1}^m \alpha_iy_i = 0, \quad 0 \leq \alpha_i \leq C ~~ \forall ~~ i = 1,...,m
\end{align}
Use of this formulation is generally referred to as Multiple Kernel Learning (MKL). This formulation is a LP in $\mu$ for fixed $\alpha$ and a QP in $\alpha$ for fixed $\mu$. Recently, a number of highly efficient two-step methods have been proposed which exploit this formulation, including SimpleMKL as in~\cite{rakotomamonjy_2008}.  These methods alternate between fixing $\mu$ and optimizing $\alpha$, then fixing $\alpha$ and optimizing $\mu$, adding the constraint that $\sum_i \mu_i=1$ using a projected gradient descent. Other two-step solvers include~\cite{gonen2011multiple}. Two-step MKL solvers typically have a significantly reduced computational complexity compared with SDP-based approaches and can typically handle thousands of data points and thousands of basis kernels.

In Section~\ref{sec:3}, we propose a parameterization of kernels using positive matrices which avoids the need for the selection of basis kernels. Moreover, we show that this parameterization can be combined with MKL algorithms directly in SimpleMKL through the use of a randomly generated basis of kernels.

\section{Positive Matrices Parameterize Positive Kernels} \label{sec:3}
To begin this section, we propose a framework for using positive matrices to parameterize positive kernels.  This is a generalization of a result initially proposed in~\cite{recht2006convex}. In Subsection~\ref{subsec:GPK} we apply this framework to obtain generalized polynomial kernels. In Subsection~\ref{subsec:TK}, we use the framework to obtain the TK class.

\begin{prop}\label{prop:kernel}
Let $N$ be any bounded measurable function $N: \mcl X \times Y \rightarrow \R^q$ on compact $\mcl X$ and $Y$ and $P \in \R^{q \times q}$ be a positive semidefinite matrix $P\geq 0$. Then
\begin{equation}
k(x,y)=\int_{\mcl X} N(z,x)^T P N(z,y) dz\label{eqn:kernel}
\end{equation}
is a positive kernel function.
\end{prop}
\begin{proof}
Since $N$ is bounded and measurable, $k(x,y)$ is bounded and measurable. Since $P \geq 0$, there exists $P^{\half}$ such that $P=(P^{\half})^TP^{\half}$. Now for any $f \in L_2[Y]$ define
\[
g(z)=  \int_{Y}  P^\half N(z,x)  f(x) dx.
\]
Then
\begin{align*}
\int_{Y} \int_Y f(x) k(x,y) f(y)dxdy &=  \int_{Y} \int_Y \int_{\mcl X} f(x)  N(z,x)^T P N(z,y)  f(y) dz dx dy \\
&=   \int_{\mcl X}  \left(\int_{Y}  P^{\half} N(z,x) f(x)dx\right)^T \left( \int_Y  N(z,y)  P^{\half} f(y) dy\right) dz \\
&= \int_{\mcl X} g(z)^T g(z) dz \ge 0.
\end{align*}
\end{proof}
For a given $N$, the map $P \mapsto k$ in Proposition~\ref{prop:kernel} is linear. Specifically,
\[
k(x,y)=\sum_{i,j} P_{i,j} G_{i,j}(x,y),
\]
where,
\[
G_{i,j}(x,y)=\int_{\mcl X}N_{i}(z,x)N_{j}(z,y)dz.
\]

\subsection{Generalized Polynomial Kernels (GPK)}\label{subsec:GPK}
Let $Y = \R^n$ and define $Z_d: \R^n \rightarrow \R^q$ to be the vector of monomials of degree $d$. If we now define $N_P(z,y)=Z_d(y)$, then $k$ as defined in Proposition~\ref{prop:kernel} is a polynomial of degree $2d$. The following result is from~\cite{peet_SICON_2009}.
\begin{lem}
A polynomial $k$ of degree $2d$ is a positive polynomial kernel if and only if there exists some $P\geq 0$ such that
\begin{equation}
k(x,y)=Z_d(x)^T P Z_d(y).\label{eqn:poly_kernel}
\end{equation}
\end{lem}
This lemma implies that a representation of the form of Equation~\eqref{eqn:kernel} is necessary and sufficient for a generalized polynomial kernel to be positive. For convenience, we denote the set of generalized polynomial kernels of degree $d$ as follows.
\begin{align}
 \mcl K_P^d &:= \{ k\;:\; k(x,y) = Z_d(x)^T P Z_d(y) \; : \; P \geq 0  \} \label{polyKernel}
 \end{align}
Unfortunately, however, polynomial kernels are never universal and hence we propose the following universal class of TK kernels, each of which is defined by polynomials, but which are not polynomial.

\subsection{Tessellated Kernels}\label{subsec:TK}
To begin, we define the indicator function for the positive orthant as
\[
I_+(z) = \begin{cases}
    1       & \quad z \ge 0\\
    0  & \quad \text{otherwise,}\\
\end{cases}
\]
where $z\ge 0$ means $z_i \ge 0$ for all $i$.
Now define $Z_d: \R^n  \times \R^n \rightarrow \R^q$ to be the vector of monomials of degree $d$ in $\R^{2n}$. We now propose the following choice of $N: \R^n \times \R^n \rightarrow \R^{2q}$.
\begin{equation}
N^d_{T}(z,x) = \bmat{Z_d(z,x)I_{+}(z-x) \\ Z_d(z,x)I_{+}(x-z) }=\begin{cases}
       \bmat{Z_d(z,x)\\0}\vspace{2mm} & z \ge x\\
       \bmat{0\\Z_d(z,x)} & x \ge z\\
       0 &\text{otherwise}
     \end{cases}  \label{eqn:N}
\end{equation}
Equipped with this definition, we define the class of Tessellated Kernels as follows.
\[
\mcl K^d_T:=\left\{k\;:\; k(x,y) = \int_{\mcl X} N_T^d(z,x)^T P N^d_{T}(z,y) dz,\; P\ge 0\right\},\qquad \mcl K_T:=\{k\;:\; k\in \mcl K_T^d,\; d \in \N\}
\]

\subsection{Representation of TK Kernels Using Polynomials}
The following result shows that any $k \in \mcl K_T$ is piecewise polynomial. Specifically, if we define the partition of $ \R^n$ into $2^n$ orthants - parameterized by $\beta \in \{0,1\}^n$ as $\{X_\beta\}_{\beta \in \{0,1\}^{n}} $ where
\begin{equation}\label{eqn:Xbeta}
X_{\beta} : = \left\{ x \in \R^n  \;:\; \substack{ x_j\ge 0  \text{ for all }  j : \beta_j=0, \\
x_i \le 0  \text{ for all }  i : \beta_i=1}\right\},
\end{equation}
then for any $k \in \mcl K_T$, there exist $k_\beta$ such that
\[
k(x,y)=\begin{cases}
         k_\beta(x,y), & \mbox{if } x-y\in X_{\beta}.
       \end{cases}  \vspace{-2mm}
\]

\begin{lem}\label{lemma1} Suppose that for $a<b \in \R^n$, $Y=\mcl X=[a,b]$, $N$ is as defined in Eqn.~\eqref{eqn:N},
\[
P = \bmat{P_{11}& P_{12}\\ P_{21}& P_{22}}> 0,
\]
$k$ is as defined in Eqn.~\eqref{eqn:kernel} and $\{X_\beta\}_{\beta \in \{0,1\}^{n}} $ is defined in Eqn.~\eqref{eqn:Xbeta}. Then
\begin{equation}
k(x,y) = \begin{cases}
         k_{\beta}(x,y) & \text{if }\; x-y \in X_\beta.
\end{cases}
\end{equation}
where the $k_\beta$ are polynomials defined as
\begin{align*}
k_\beta(x,y)&= \int\limits_{\beta_1y_1 + (1-\beta_1)x_1}^{b_1} \dots \int\limits_{\beta_ny_n + (1-\beta_n)x_n}^{b_n}  Z_d(z,x)^T Q_1 Z_d(z,y) dz +k_0(x,y),
\end{align*}
where
\begin{align*}
k_0(x,y)&= \hspace{-1mm}\int_{x}^b  \hspace{-2mm} Z_d(z,x)^T\hspace{-1mm}Q_2 Z_d(z,y) dz + \int_{y}^b \hspace{-2mm} Z_d(z,x)^T Q_3 Z_d(z,y) dz + \int_{a}^b \hspace{-2mm} Z_d(z,x)^T P_{22} Z_d(z,y) dz,
\end{align*}
and
\[
Q_1 \hspace{-1mm}=\hspace{-1mm}P_{11}-P_{12}-P_{21}+P_{22}, \quad Q_2\hspace{-1mm}=\hspace{-1mm}P_{12}-P_{22},\quad Q_3\hspace{-1mm}=\hspace{-1mm}P_{21}-P_{22}.
\]
\end{lem}
\begin{proof}

Given $N$ as defined above, if we partition $P=\bmat{P_{11}& P_{12}\\ P_{21}& P_{22}}$ into equal-sized blocks, we have \vspace{-3mm}
\begin{align*}
k(x,y)=\int_{\mcl X} N(z,x)^T P N(z,y) dz =\sum_{i,j=1}^2  \int_{(x,y,z) \in {\mcl X}_{ij}} Z_d(z,x)^T P_{i,j}Z_d(z,y) dz
\end{align*}
where\vspace{-3mm}
\begin{align*}
{\mcl X}_{ij} &:=  \{ (x,y,z) \in \R^{3n} \;:\;  I_{+}((-1)^{j}(z-x))I_{+}((-1)^{j}(z-y)) = 1  \}.
\vspace{-3mm} \end{align*}
From the definition of $\mcl X_{ij}$ we have that,
\begin{align*}
{\mcl X}_{11} =& \{z\in {\mcl X} \; : \; z_i \ge p^*_i(x,y),\; i=1,\cdots,n \} \\
{\mcl X}_{12} =&  \{z \in {\mcl X} \;:\; z_i \ge x_i,\; i=1,\cdots,n \} / {\mcl X}_{11}\\
{\mcl X}_{21} =&  \{z \in {\mcl X} \;:\; z_i \ge y_i,\; i=1,\cdots,n \} / {\mcl X}_{11} \\
{\mcl X}_{22} =&   {\mcl X}  / \left({\mcl X}_{11}\cup {\mcl X}_{12} \cup {\mcl X}_{21}\right).
\end{align*}
where $p_i^*(x,y)=\max\{x_i,y_i\}$ and $p_i^*(x,y) = \beta_iy_i + (1-\beta_i)x_i$.
By the definitions of ${\mcl X}_{11},{\mcl X}_{12},{\mcl X}_{21},$ and ${\mcl X}_{22}$ we have that,
\begin{align} \label{kContinuous}
k(x,y)=&\hspace{-1mm}\int_{p^*(x,y)}^b \hspace{-9mm}Z_d(z,x)^T \hspace{-1mm}\left(P_{11}-P_{12}-P_{21}+P_{22}\right) Z_d(z,y) dz \nonumber + \hspace{-1mm}\int_{x}^b \hspace{-2mm} Z_d(z,x)^T\hspace{-1mm} \left(P_{12}-P_{22}\right) Z_d(z,y) dz \nonumber\\
&+ \int_{y}^b \hspace{-2mm} Z_d(z,x)^T \left(P_{21}-P_{22} \right)Z_d(z,y) dz + \int_{a}^b \hspace{-2mm} Z_d(z,x)^T P_{22} Z_d(z,y) dz. \end{align}  \end{proof}

Note that the number of domains $X_\beta$ used to define the piecewise polynomial $k$ is $2^n$, which does not depend on $q$ (the dimension of $P_{ij}$). Thus, even if $Z_d=1$, the resulting kernel is partitioned into $2^n$ domains. The size of $Z_d(x,y) \in \R^q$ only influences the degree of the polynomial defined on each domain.

The significance of the partition does not lie in the number of domains of the kernel, however. Rather, the significance of the partition lies in the resulting classifier, which, for a given set of training data $\{x_i\}_{i=1}^m$, has a domain tessellated into $(m+1)^n$ tiles, $X_\gamma$, where $\gamma \in \{0,\cdots,m\}^n$.  Although the training data is unordered, we create an ordering using $\Gamma(i,j): \{0,\cdots,m\}\times \{1,n\} \rightarrow \{1,\cdots,m\}$ where $\Gamma(i,j)$ indicates that among the $j$th elements of the training data, $x_{\Gamma(i,j)}$ has the $i$th largest value. That is,
\[
[x_{\Gamma(i-1,j)}]_j\le [x_{\Gamma(i,j)}]_j\le [x_{\Gamma(i+1,j)}]_j \qquad \forall i=1,\cdots,m-1,\quad j =1,\cdots,n.
\]
Now, for any $\gamma \in \{0,\cdots,m\}^n$, we may define an associated tile
\[
X_{\gamma}:=\left\{z\;:\;[x_{\Gamma(\gamma_j,j)}]_j\le z_j\le [x_{\Gamma(\gamma_j+1,j)}]_j,\; j=1,\cdots,n\right\}.
\]
The classifier may now be represented as
\begin{align*}
f(z) &= \sum_{i=1}^m \alpha_iy_i k(x_i,z) + b\\
&=
         f_{\gamma}(z) \;\; \forall z \in X_{\gamma}.
\end{align*}
To define the $f_\gamma$, we associate with every tile $\gamma$ and datum $i$ an orthant $\beta(i,\gamma)$ which denotes the position of tile $T_\gamma$ relative to datum $x_i$ - i.e. $T_\gamma$ is in the orthant $\beta(i,\gamma)$ centered at the point $x_i$. Specifically,
\[
\beta(i,\gamma)_j=\begin{cases}0&[x_{\Gamma(\gamma_j,j)}]_j \ge [x_i]_j \\1&\text{otherwise.}\end{cases}
\]
Now we may define
\[
f_{\gamma}(z)= \sum_{i=1}^m \alpha_i y_i k_{\beta(i,\gamma)}(x_i,z)
\]
which is a polynomial for every $\gamma$.
%
%
%
%
In this way, each data point further divides the domains which it intersects, resulting in $(m+1)^n$ disjoint sub-domains, each with associated polynomial classifier.

Thus we see that the number of domains of definition of the classifier grows quickly in $m$, the number of training data points.  For instance, with n = 2 there are 100 tiles for just 9 data points. This growth is what makes TK kernels universal - as will be seen in Section IV.

\begin{figure}[t]
    \centering
    \begin{subfigure}[t]{0.5\textwidth}
        \centering
\includegraphics[width=\textwidth]{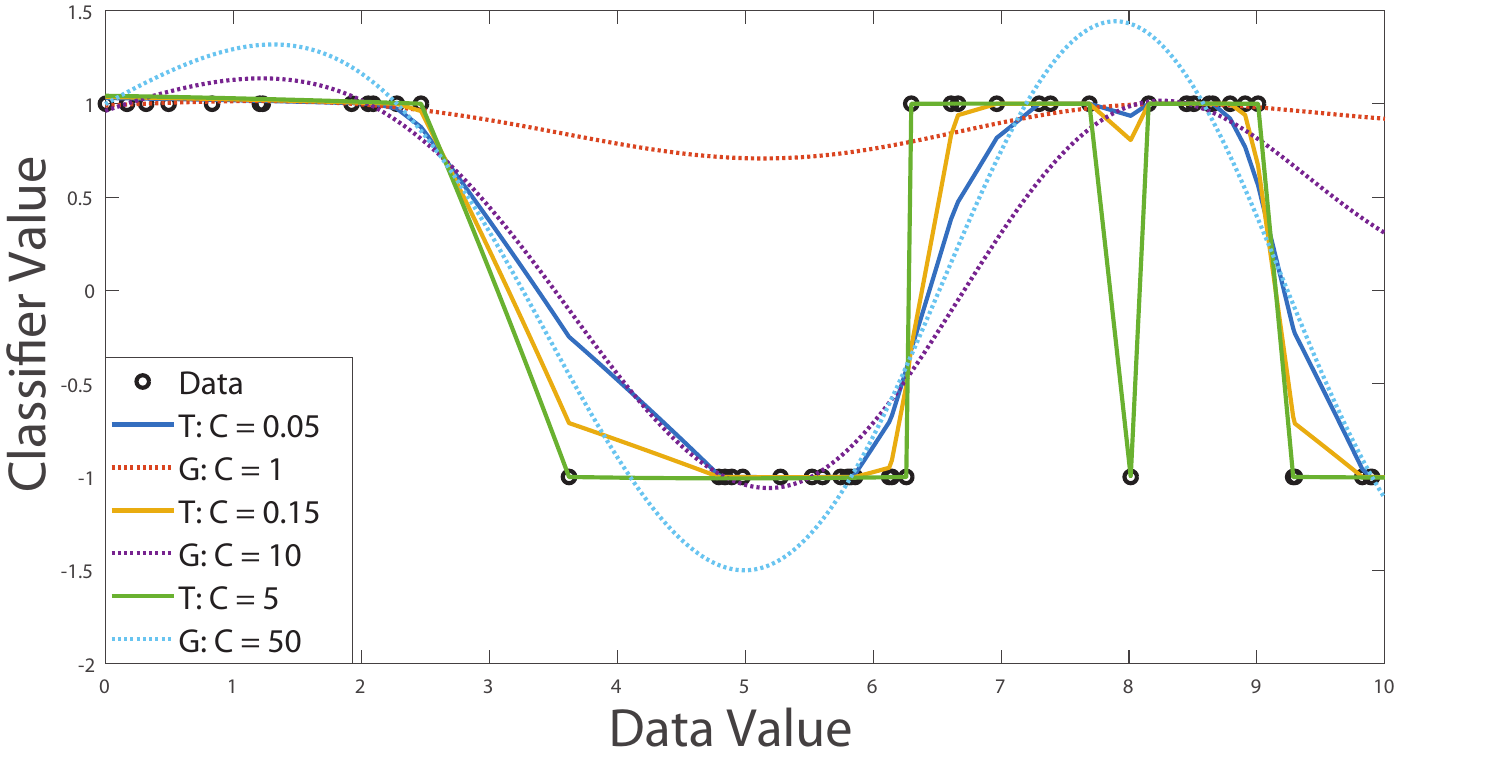}
        \caption{Optimal classifier, $f(z)$ for labelling a 1 dimensional data set using a degree one TK (solid lines), and a positive combination of Gaussian kernels (dotted lines) with three different penalty weights $C$.}
    \end{subfigure}%
    ~
    \begin{subfigure}[t]{0.5\textwidth}
        \centering
\includegraphics[width=\textwidth]{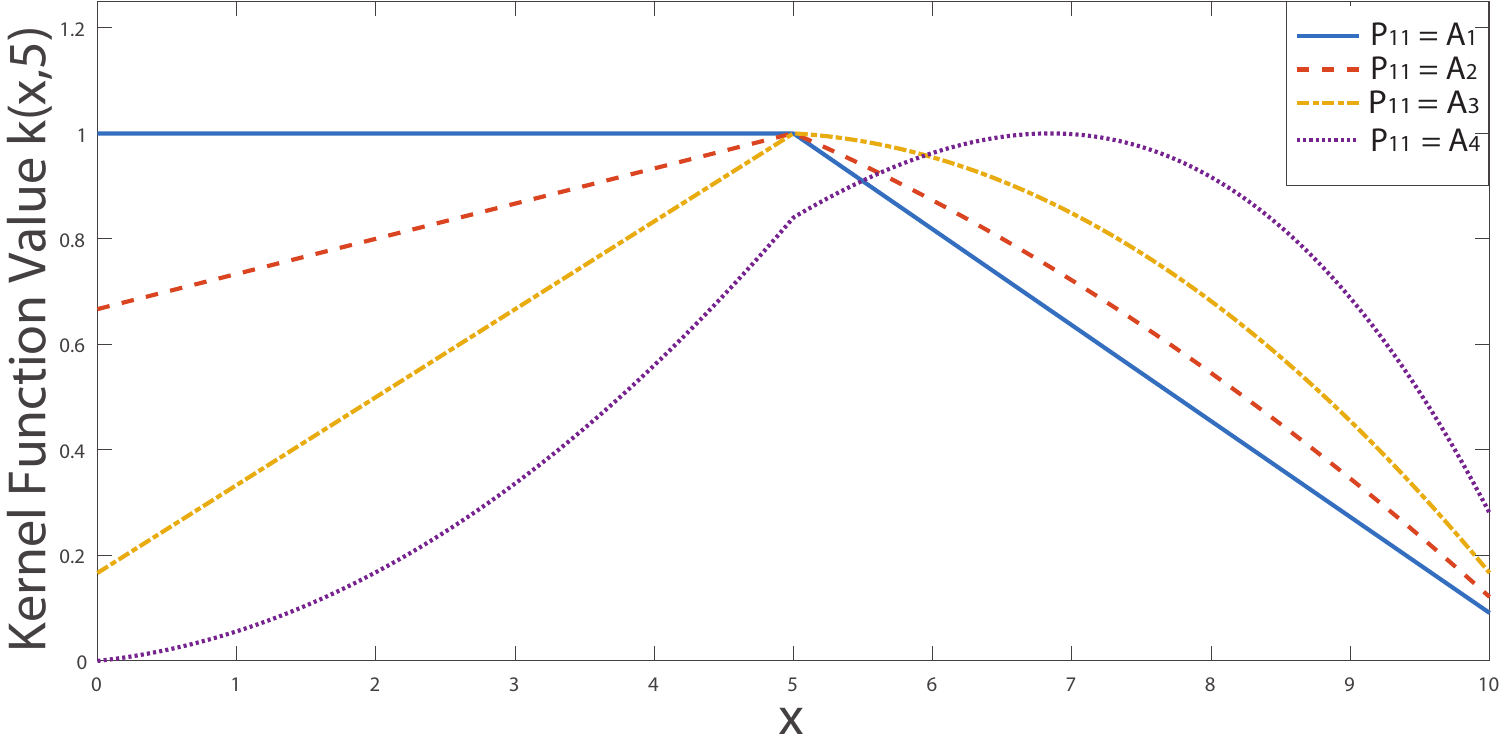}
        \caption{Normalized kernel function $k(5,z)$  using $P_{1,1}=A_i$ from \eqref{ExampleMatrix} and $P_{1,2} = P_{2,1} = P_{2,2} = 0$.}
    \end{subfigure}
    \caption{This figure depicts the optimal classifier for labelling a 1-dimensional data set compared to Gaussian classifiers as well as the normalized kernel function, $k(5,z)$,  using different $P_{11}$ matrices matrices and $\mcl X=[0,10]$.}\label{fig:KernelFunctionBasis} \vspace{-5mm}
\end{figure}

In Figure \ref{fig:KernelFunctionBasis}(a) we see the function, $f(z) = \sum_{i=1}^m \alpha_iy_i k(x_i,z) + b$, for a degree 1 TK kernel trained for a 1-dimensional labeling problem as compared with a Gaussian kernel. We see that the TK classifier is continuous, and captures the shape of the generator better than the Gaussian. Note that the TK classifier is not continuously differentiable and the derivative can change precipitously at the edges of the tiles. However, if we decrease the inverse regularity weight $C$ in the objective function of Optimization Problem~\eqref{eqn:SVM}, then this has the effect of smoothing the resulting classifier. In Figure \ref{fig:KernelFunctionBasis}(a), as $C$ decreases we see that the changes in slope at edges of the tiles decrease.

To illustrate that the function $k(x_i,z)$ is a piecewise polynomial tessellated by the training datum, we plot the value of an assortment of TK kernels in one dimension in Figure~\ref{fig:KernelFunctionBasis}(b).  We use training datum $x_i = 5$, and a selection of different positive matrices where $P_{1,2} = P_{2,1} = P_{2,2} = 0$ and $P_{1,1}=A_i$ for $i=1,\ldots,4$ where
\begin{align} \label{ExampleMatrix}
A_1 = \bmat{1&0\\0&0}, \quad A_2 = \bmat{1&0\\0&.1}, \quad A_3 = \bmat{1&0\\0&1}, A_4 = \bmat{0&0&0\\0&1&0\\0&0&1}.
\end{align}
In the first three cases the monomial basis is of degree 1, while in the fourth case the monomial basis is of degree 2 - for simplicity we exclude monomials with $z$.  These different matrices all illustrate changes in slope which occur at the training datum. 

\section{Properties of the Tessellated Class of Kernel Functions} \label{sec:4} \vspace{-3mm}
In this section, we prove that all TK kernels are continuous and universal and that the TK class is pointwise dense in all kernels.

\subsection{TK Kernels Are Continuous}
Let us begin by recalling that for any $P\geq 0$ and $N(z,x)$,
\[
k(x,y)=\int_{\mcl X} N(z,x)^T P N(z,y) dz
\]
is a positive kernel and recall that for the TK kernels, we have
\[
N(z,x) = \begin{bmatrix}
       Z_d(z,x)I_{+}(z-x) \\[0.3em]
       Z_d(z,x)I_{+}(x-z)
     \end{bmatrix}.
\]
By the representer theorem this implies that the classifiers consist of functions of the form

\[
f(y) = \sum_{i=1}^m \alpha_i\int_{\mcl X} N(x_i,z)^TPN(y,z) dz.
\]

The following theorem establishes that such functions are necessarily continuous.

\begin{thm} \label{thm:continuity}
Suppose that for $a<b \in \R^n$, $Y={\mcl X}=[a,b]$, $P\geq 0$, $N$ is as defined in Eqn.~\eqref{eqn:N}
for some $d \ge 0$ and $k$ is as defined in Eqn.~\eqref{eqn:kernel}. Then $k$ is continuous and for any $\{x_i\}_{i=1}^m$ and $\alpha \in \R^m$, the function \vspace{-3mm}
\[
f(z) = \sum_{i=1}^m \alpha_i k(x_i,z), \vspace{-3mm}
\]
is continuous.
\end{thm}

\begin{proof}
Partition $P$ as follows
\[
P = \bmat{P_{11}& P_{12}\\ P_{21}& P_{22}} > 0.
\]

To prove that $f(z)$ is continuous we need only prove that $k(x,y)$ is continuous.  Applying Lemma 3 we may define $k(x,y)$ as
\begin{equation}
k(x,y) = \begin{cases}
         k_{\beta}(x,y) & \text{if }\; x-y \in X_\beta.
\end{cases}
\end{equation}
where the $k_\beta$ are polynomials defined as
\begin{align*}
k_\beta(x,y)&= \int_{\theta_{\beta,1}(x,y)}^{b_1} \dots \int_{\theta_{\beta,n}(x,y)}^{b_n}   Z_d(z,x)^T Q_1 Z_d(z,y) dz +k_0(x,y),
\end{align*}
where $x-y \in X_\beta$, $\theta_{\beta,i}(x,y) = \beta_i y_i + (1-\beta_i)x_i$, $Q_1 =P_{11}-P_{12}-P_{21}+P_{22}$, and $k_0(x,y)$ is a polynomial.
To expand $k_\beta(x,y)$, we use multinomial notation for the monomials in $Z_d$. Specifically, we index the elements of $Z_d$ as $Z_{d}(z,x)_i=z^{\gamma_{i}}x^{\delta_{i}}$ where $\gamma_{i},\delta_{i} \in \N^n$ for $i=1,\cdots,q$ and where therefore $z^{\gamma_{i}}x^{\delta_{i}} = \prod_{j=1}^n z_j^{\gamma_{i,j}} x_j^{\delta_{i,j}}$.  Then \vspace{-3mm}
\begin{align}
\int\limits_{\theta_{\beta,1}(x,y)}^{b_1} \dots \int\limits_{\theta_{\beta,n}(x,y)}^{b_n}  Z_d(z,x)^T Q_1 Z_d(z,y) dz &=\int\limits_{\theta_{\beta,1}(x,y)}^{b_1} \dots \int\limits_{\theta_{\beta,n}(x,y)}^{b_n}   \sum_{k,l}\left(Q_1\right)_{k,l} x^{\delta_{k}}z^{\gamma_{k}}z^{\gamma_{l}}y^{\delta_{l}} dz\notag \\
&=\sum_{k,l}\left(Q_1\right)_{k,l}   x^{\delta_{k}}y^{\delta_{l}} \int\limits_{\theta_{\beta,1}(x,y)}^{b_1} \dots \int\limits_{\theta_{\beta,n}(x,y)}^{b_n}  \hspace{-5mm} z^{\gamma_{k}+\gamma_{l}} dz.\label{multinomial}
\end{align}
Expanding the integrals in~\eqref{multinomial}, each has the form
\begin{align*}
\int_{\theta_{\beta,1}(x,y)}^{b_1} \dots \int_{\theta_{\beta,n}(x,y)}^{b_n}   z^{\gamma_{k}+\gamma_{l}} \;dz & = \prod_{j=1}^n \hspace{2mm}  \int_{\theta_{\beta,j}(x,y)}^{b_j} z_j^{\gamma_{k,j}+\gamma_{l,j}} ~ dz_j \\
& = \prod_{j=1}^n \hspace{2mm} \frac{z_j^{\gamma_{k,j}+\gamma_{l,j}+1}}{\gamma_{k,j}+\gamma_{l,j}+1} \Big|^{b_j}_{\theta_{\beta,j}(x,y)} \\
& = \prod_{j=1}^n \hspace{2mm} \frac{b_j^{\gamma_{k,j}+\gamma_{l,j}+1}}{\gamma_{k,j}+\gamma_{l,j}+1} -\frac{\theta_{\beta,j}(x,y)^{\gamma_{k,j}+\gamma_{l,j}+1}}{\gamma_{k,j}+\gamma_{l,j}+1}. \\
\end{align*}
Since $\theta_{\beta,j}(x,y)$ is equivalent to $\max(x_j,y_j)$, and can be written as the continuous function,
\[
\theta_{\beta,j}(x,y) = \frac{1}{2}(x_j+y_j+|x_j-y_j|),
\]
we conclude that $k(x,y)$ is the product and summation of continuous functions and therefore $k$ and the resulting classifiers are both continuous.
\end{proof}

\subsection{TK Kernels Are Universal} In addition to continuity, we show that any TK kernel with $P> 0$ has the universal property.  Recall the following definition of universality.

\begin{defn}
A kernel $k:\mcl X \times \mcl X \rightarrow \R$ is said to be \textit{universal} on the compact metric space $\mcl X$ if it is continuous and there exists an inner-product space $\mcl W$ and feature map, $\Phi : \mcl X \rightarrow \mcl W$ such that $k(x,y)=\ip{\Phi(x)}{\Phi(y)}_{\mcl W}$ and where the unique Reproducing Kernel Hilbert Space (RKHS),
\[
\mcl H:=\{f\;:\;f(x)=\ip{v}{\Phi(x)},\; v \in \mcl W\}
\]
with associated norm $\norm{f}_{\mcl H} :=\inf_{v}\{\norm{v}_{\mcl W}\;:\; f(x)=\ip{v}{\Phi(x)}\}$ is dense in $\mcl C(\mcl X):=\{f \,: \, \mcl X \rightarrow \R \;:\; $f$\, \text{ is continuous}\}$ where $\norm{f}_{\mcl C}:=\sup_{x\in X} |f(x)|$.
\end{defn}
The following theorem shows that any TK kernel with $P > 0$ is necessarily universal.

\begin{thm} \label{thm:universal}
Suppose $k$ is as defined in Eqn.~\eqref{eqn:kernel} for some $P>0$, $d \in\N$ and $N$ as defined in Eqn.~\eqref{eqn:N}. Then $k$ is universal for $Y= {\mcl X}=[a,b]$, $a<b \in \R^n$.
\end{thm}
\begin{proof}
Without loss of generality, we assume $Y={\mcl X}=[0,1]^n$. If $P>0$, then there exist $\epsilon_i$ such that $P=P_0+ \sum_i \epsilon_i P_i$ where $P_0 > 0$ and
\[
P_1=\bmat{0&0\\0&1} \otimes \bmat{e_1,0,\dots,0}
\]
where $\{e_1\}$ is the first canonical basis of $\R^n$.  In this case
\begin{align*}
k(x,y) &=\hat{k}(x,y)+ \underbrace{\prod_{i=1}^n \epsilon_i  \text{min} \{ x_i, y_i \}}_{k_1(x,y)},
\end{align*}
where $\hat{k}$ is a positive kernel. Since the hypothesis space satisfies the additive property (See~\cite{wang2013} and~\cite{borgwardt2006}), if $k_1$ is a universal kernel, then $k$ is a universal kernel.

Recall that for a given kernel, the hypothesis space, $\mcl H$, can be characterized as the closure of $\text{span}\{k(y,\cdot)\,:\,y \in \mcl X \}$.  Now, consider
\[
\text{span}\{k_1(y,\cdot)\,:\,y \in \mcl X \},
\]
which consists of all functions of the form
\[
f(x)=\sum_j c_j\prod_{i=1}^n \min\{[y_j]_i,x_i\}.
\]
Now
\[
\text{min} \{[y_j]_i,x_i \}=\begin{cases}
                                 x_i, & \mbox{if } x_i \le [y_j]_i \\
                                 [y_j]_i, & \mbox{otherwise}.
                               \end{cases}
\]
For $n=1$, we may construct a triangle function of height 1 centered at $y_2$ as
\[f(x)= \sum_{i=1}^3 \frac{\alpha_i}{\epsilon} k_1(y_i,x) =\begin{cases}0, & \mbox{if } x < y_1 \\
                                 \delta (x-y_1), & \mbox{if }y_1 \leq x < y_2 \\
                                 1-\delta (x-y_2), & \mbox{if } y_2 \leq x < y_3 \\
                                 0, & \mbox{if } y_3 < x,
                               \end{cases} \]
where $\delta = y_1-y_2 = y_2-y_3$, and
\[
\alpha_1 = -\delta, \quad \alpha_2 = 2\delta, \quad \alpha_3 = -\delta.
\]
By taking the product of triangle functions in each dimension, we obtain the pyramid functions which are known to be dense in the space of continuous functions on a compact domain (See~\cite{shekhtman1982piecewise}). We conclude that $k_1$ is a universal kernel and hence $k$ is universal.
\end{proof}

This theorem implies that even if the degree of the polynomials is small, the kernel is still universal. Specifically, in the case when $n=1$ and $d=0$, the set $\mcl K_T^0$ is universal yet contains only three parameters (the elements of the symmetric $P\in \R^{2\times 2}$).

\subsection{TK Kernels Are Pointwise Dense in All Kernels}
In the previous two subsections, we have shown that TK kernels are continuous and
universal. Furthermore, as shown in Section~\ref{sec:3}, the TK class admits
a linear parameterization. The remaining question, then, is whether TK
kernels are superior in some \textit{performance} metric to other classes of
universal kernels such as Gaussian kernels. First, note that the universal
property is of the kernel itself and which is extended to a class of kernels
by requiring all kernels in that class to satisfy the property. However,
although a kernel may be universal, it may not be well-suited to SVM.
Expanding on this point, although it is known that any universal kernel may
be used to separate a given set of data, it can be shown that for any given set of normalized data, $\{x_i,y_i\}$, there exists a universal kernel, $k$, for which the solution to Optimization Problems~\eqref{eqn:SVM} and~\eqref{KernelSVM}
is arbitrarily suboptimal - e.g. by increasing the bandwidth of the Gaussian kernel.

To address the question of performance, we propose the \textit{pointwise density} property. This property is defined on a set of kernels and guarantees that there is some kernel in the set of kernels for which the solution to Optimization Problems~\eqref{eqn:SVM} and~\eqref{KernelSVM} is optimal. Specifically, we have the following.
\begin{defn}
The set of kernels $\mcl K$ is said to be \textbf{pointwise dense} if for any positive kernel, $k^*$, any set of data $\{x_i\}_{i=1}^m$, and any $\epsilon>0$, there exists $k \in \mcl K$ such that $\norm{k(x_i,x_j)-k^*(x_i,x_j)}\le \epsilon$.
\end{defn}
This definition implies that a set of kernels can approximate any given positive kernel arbitrarily well. To illustrate the importance of the pointwise density property, in Subsubsection~\ref{subsubsection:positivity} we show that for a large class of kernel learning problems, the value of the optimal kernel is not pointwise positive - i.e. $k(x,y)\not\ge 0$ for all $x,y\in \mcl X$. This is significant because almost all commonly used kernels are pointwise non-negative. Indeed we find that the elements of the optimal kernel matrix are negative as frequently as they are positive.

\subsubsection{Optimal Kernels Are Not Pointwise Positive}\label{subsubsection:positivity} To demonstrate the necessity of negative values in optimal kernel matrices, we analytically solve the following SDP derived from Optimization Problem~\eqref{SDP} which determines the optimal kernel matrix ($K^*$) given the labels $y$ of a problem and a ``penalty'' parameter $C$, but with no constraint on the form of the kernel function (other than it be PD).  \vspace{-2mm}

\begin{align}\label{OPTKernel}
&\underset{ t \in \R, K \in \R^{m \times m}, \gamma \in \R,\nu \in \R^{m}, \delta \in \R^m}{\text{min}} \quad t, \\ \nonumber
& \text{subject to:}   \begin{pmatrix}
  G & e + \nu - \delta + \gamma y \\ \nonumber
  (e +  \nu- \delta + \gamma y)^T & t-C \delta^T e
 \end{pmatrix} \geq 0\\ \nonumber
 &  \nu \geq 0, \qquad \delta \geq 0, \qquad K \geq 0, \qquad \text{trace}( K) = m, \qquad G_{i,j} = y_i K_{i,j} y_j \\ \nonumber
\end{align}
The following theorem finds an analytic solution of this optimization problem.
\begin{thm} \label{thm:OptimalKernel}
Let $y_i \in \{ 1,-1 \}$ for $i = 1, \cdots, m$ and $C \geq \frac{2}{m}$, then the solution to Optimization Problem~\ref{OPTKernel} is,
\[ \nu^* = 0, \quad \gamma^* = -\sum_{i=1}^m \frac{y_i}{m}, \quad \delta^* = 0, \quad t^* =  \frac{\norm{e-\gamma^* y}_2}{m},\]
and
$K^* = \frac{m}{\norm{e + \gamma^* y}_2^2} {\mcl Y}( e + \gamma^* y)(e + \gamma^* y)^T {\mcl Y}$ where ${\mcl Y} = \text{diag}(y) $.
\end{thm}
\begin{proof}
We first show that $K^*= U\Sigma U^T$, where
\[
U = {\mcl Y}\bmat{ \frac{(e+\nu^*-\delta^*+\gamma^* y)}{\norm{(e+\nu^*-\delta^*+\gamma^* y)}_2} & \cdots },  \Sigma = \bmat{ m &  0 \\ \ 0  & 0}.
\]

Optimization Problem~\eqref{OPTKernel} is equivalent to
\begin{align}
&\underset{  K \in \R^{m \times m}, \gamma \in \R,\nu \in \R^{m}, \delta \in \R^m}{\text{min}}
\quad (e+\nu-\delta+\gamma y)^T ({\mcl Y}K{\mcl Y})^{-1}(e+\nu-\delta+\gamma y) + 2C\delta^T e \\ \nonumber
& \text{subject to:}
\quad  \nu \geq 0, \qquad \delta \geq 0, \qquad K \geq 0, \qquad \text{trace}(K) = m. \nonumber
\end{align}
This problem can be separated into subproblems as
\begin{align}
&\min_{\substack{  \gamma \in \R, \nu \in \R^{m}, \delta \in \R^m\\  \nu \geq 0,\; \delta \geq 0}} \qquad  \min_{\substack{  K \in \R^{m \times m},\\  K \geq 0,  \;\text{trace}(K) = m }}
\quad (e+\nu-\delta+\gamma y)^T ({\mcl Y}K{\mcl Y})^{-1}(e+\nu-\delta+\gamma y) + 2C\delta^T e. \nonumber
\end{align}

Now, for any feasible $K$, we have that $K\ge 0$ and $\bar \sigma(K) \le m$ and hence
\begin{align*}
(e+\nu-\delta+\gamma y)^T ({\mcl Y}K{\mcl Y})^{-1}(e+\nu-\delta+\gamma y) & \geq \frac{1}{\bar{\sigma}(K)}\norm{e+\nu-\delta+\gamma y}^2_2\\
& \geq \frac{1}{m}\norm{e+\nu-\delta+\gamma y}^2_2.
\end{align*}

Now, we propose $K = U\Sigma U^T$ and show that it is optimal, where
\[
U = {\mcl Y}\bmat{ \frac{(e+\nu^*-\delta^*+\gamma^* y)}{\norm{(e+\nu^*-\delta^*+\gamma^* y)}_2} & V }, \quad \Sigma = \bmat{ m &  0 \\ \ 0  & 0}.
\]
and $V$ is any unitary completion of the matrix $U$. Then $K \ge 0$, $\text{trace}(K)=m$, and
\begin{align*}
&(e+\nu-\delta+\gamma y)^T \left( {\mcl Y} K {\mcl Y} \right)^{-1} \hspace{-1mm}(e+\nu-\delta+\gamma y)\\
&=(e+\nu-\delta+\gamma y)^T \left(\bmat{ \frac{(e+\nu-\delta+\gamma y)}{\norm{(e+\nu-\delta+\gamma y)}_2} & \hspace{-3mm} V } \bmat{ m &  0 \\ \ 0  & 0} \bmat{ \frac{(e+\nu-\delta+\gamma y)}{\norm{(e+\nu-\delta+\gamma y)}_2} & \hspace{-3mm} V }^T \right)^{-1} \hspace{-3mm} (e+\nu-\delta+\gamma y)\\
&=  \frac{\norm{(e+\nu-\delta+\gamma y)}^2_2}{m}.
\end{align*}
We conclude that this $K$ solves the first sub-problem and hence Optimization Problem~\eqref{OPTKernel} reduces to
\begin{align}
&\min_{\substack{   \gamma \in \R, \nu \in \R^{m}, \delta \in \R^m\\ \nu \geq 0,\; \delta \geq 0,}}
\quad \frac{\norm{(e+\nu-\delta+\gamma y)}^2_2}{m} + 2C\delta^T e. \label{OPTreduced}
\end{align}
Now let $\nu^*, \delta^*, \gamma^*$ be as defined in the theorem statement. For the convex objective
\[
f(\delta,\nu,\gamma)=\frac{\norm{e+\nu-\delta+\gamma y}_2^2}{m} + 2C\delta^T e
\]
let $\bar{y} = \frac{1}{m}\sum_{i=1}^m y_i$ and we have that
\begin{align*}
\frac{\partial f}{\partial \nu_i}(\nu^*, \delta^*, \gamma^*) &= \frac{2+2\bar{y}y_i}{m} \geq \frac{2 + -2(1)(1)}{m} \geq 0,
\end{align*}
and for $C\geq \frac{2}{m}$
\begin{align*}
\frac{\partial f}{\partial \delta_i}(\nu^*, \delta^*, \gamma^*) &= \frac{2mC-2-2\bar{y}y_i}{m} \geq \frac{4-2-2(1)(1)}{m} \geq 0.
\end{align*}
Finally
\begin{align*}
\frac{\partial f}{\partial \gamma}(\nu^*, \delta^*, \gamma^*)= \frac{1}{m} \sum_{i=1}^m 2y_i - 2\bar{y}=  -2\bar{y}+2\frac{1}{m}\sum_{i=1}^m y_i= 0.
\end{align*}
Hence the KKT conditions are satisfied and since the optimization problem is convex, $(\nu^*, \delta^*, \gamma^*)$ is optimal.
\end{proof}

This result shows that for binary labels, the optimal kernel matrix has an analytic solution. Furthermore, if  we consider the case where $\sum_{i=1}^m y_i = 0$, then $\lambda^* = 0$ and hence $K^*=yy^T$ and $K^*_{i,j} = y_iy_j$. This implies that the optimal kernel matrix consists of an equal number of positive and negative entries - meaning that kernels functions with globally positive values will not be able to approximate the optimal kernel matrix well. Furthermore, for values of $C$ less than $\frac{2}{m}$, we find numerically that the same kernel matrix is still optimal - only the values of $\delta^*$ and $\gamma^*$ are different.

\subsubsection{The GPK and TK Classes Are Pointwise Dense in All Kernels} \label{subsubsection:density}
Having demonstrated the significance of pointwise density, we now establish that both the GPK and TK kernel sets satisfy this property. For this subsubsection, we relax the strict positivity constraint $P>0$ in the definition of the TK class. In this case, the GPK class becomes a subset of the TK class. We prove pointwise density of the GPK class - a property which is then inherited by the TK class. The following lemma shows that the GPK class is a subset of the TK class.
%

\begin{lem} \label{lem:TKPoly}
  $\mcl K_P^d \subset \mcl K^d_T$
\end{lem}
\begin{proof}
If $k_p \in \mcl K_P^d$, there exists a $P_1\ge 0$ such that  $k_p(x,y) = Z_d(x)^T P_1 Z_d(y)$. Now let $J$ be the matrix such that $J Z_d(z,x)=Z_d(x)$ and define
\[P =\frac{1}{\prod_{j=1}^n \left( b_j-a_j \right) } \bmat{ J^TP_{1}J & J^TP_{1}J \\ J^TP_{1}J & J^TP_{1}J}\ge 0.
\]
Now let $k$ be as defined in Equation~\eqref{eqn:kernel}. Then $k \in \mcl K^d_T$ and
\begin{align}
k(x,y)=&\frac{1}{\prod_{j=1}^n \left( b_j-a_j \right) }\int_{p^*(x,y)}^b Z_d(z,x)^T J^T\left(P_{1}-P_{1}-P_{1}+P_{1}\right) JZ_d(z,y) dz \nonumber\\
& + \frac{1}{\prod_{j=1}^n \left( b_j-a_j \right) }\int_{x}^b  Z_d(z,x)^T J^T\left(P_{1}-P_{1}\right)J Z_d(z,y) dz \nonumber\\
&+ \frac{1}{\prod_{j=1}^n \left( b_j-a_j \right) } \int_{y}^b  Z_d(z,x)^T J^T\left(P_{1}-P_{1} \right)J Z_d(z,y) dz \nonumber \\
& + \frac{1}{\prod_{j=1}^n \left( b_j-a_j \right) } \int_{a}^b  Z_d(z,x)^T J^T P_{1} JZ_d(z,y) dz \nonumber \\
=& \frac{1}{\prod_{j=1}^n \left( b_j-a_j \right) }\int_{a}^b  Z_d(x)^T P_{1} Z_d(y) dz  \nonumber \\
=& k_p(x,y). \nonumber
\end{align}
We conclude that $k_p=k\in \mcl K_T^d$.
\end{proof}


We now use polynomial interpolation to prove that GPK kernels are pointwise dense.

\begin{thm} \label{thm:TessellatedOptimal}
For any kernel matrix $K^*$ and any finite set $\{x_i\}_{i=1}^m$, there exists a $d \in\N$ and $k\in\mcl K_P^d$ such that if $K_{i,j} = k(x_i,x_j)$, then $K=K^*$.
\end{thm}
\begin{proof}
  Since $K^*\ge 0$, $K^*=M^TM$ for some $M$.
  Using multivariate polynomial interpolation (as in~\cite{gasca_2001}), for sufficiently large $d$, we may choose $Q$ such that
  \[
  Q\bmat{Z_d(x_1)&\cdots&Z_d(x_m)}=M.
  \]
Now let
  \[
  k(x,y)=Z_d(x)^T P Z_d(y)
  \]
  where $P=Q^TQ\ge 0$. Now partition $M$ as
  \[
  M=\bmat{m_1&\cdots&m_m}.
  \]
Then $QZ_d(x_i)=m_i$ and hence
  \begin{equation*}
  \begin{split}
  K_{ij}&=Z_d(x_i)^T Q^TQ Z_d(x_j)\\
  &=m_i^Tm_j\\
  &=K_{ij}^*.
  \end{split}
  \end{equation*}
\end{proof}

\subsubsection{GPK and TK Kernels Converge Quickly to the Optimal Kernel}
In Subsubsection~\ref{subsubsection:positivity}, we obtained an analytical solution to the optimal kernel matrix. In Subsubsection~\ref{subsubsection:density}, we used polynomial interpolation to prove that GPK and TK kernels are pointwise dense in the set of all kernels. However, the degree of the polynomials used in this proof increases with the number of interpolation points. In this subsubsection, we show that in practice, a degree of only 4 or 5 can be sufficient to approximate the optimal kernel matrix with minimal error using either GPK and TK kernels.

Specifically, we consider the problem of approximating the optimal kernel matrix for a given set of data $\{x_i\}$ and given set of kernels, $\mcl K$, using both the element-wise matrix $\norm{\cdot}_1$ and $\norm{\cdot}_\infty$ norms.
\begin{align}
\min_{k\in \mcl K} \frac{\norm{K-K^*}_1}{n^2}~~~ s.t. ~~K_{i,j} = k(x_i,x_j) \label{opt:1norm} \\
\min_{k \in \mcl K} \norm{K-K^*}_\infty~~~ s.t. ~~K_{i,j} = k(x_i,x_j) \label{opt:infnorm}
\end{align}
The sets of kernels functions we consider are: $\mcl K_G^\gamma$ - the sum of $K$ Gaussians with bandwiths $\gamma_i$; $\mcl K_P^d$ - the GPKs of degree $d$; and $\mcl K_T^d$  - the TK kernels of degree $d$. That is, we choose $\mcl K\in \{\mcl K^\gamma_G,~\mcl K^d,~\mcl K^d_T\}$ where for convenience, we define the class of sums of Gaussian kernels of bandwidths $\gamma\in \R^{K}$ as follows.
\begin{align}
 \mcl K^\gamma_G &:= \left\{k\;:\; k(x,y) = \sum_{i = 1}^K \mu_i \text{e}^\frac{||x-y||_2^2}{\gamma_i} \; : \; \mu_i > 0  \right\}\label{eqn:GK}
 \end{align}

\begin{figure}[t]
    \centering
    \begin{subfigure}[t]{0.5\textwidth}
        \centering
\includegraphics[width=.91\textwidth]{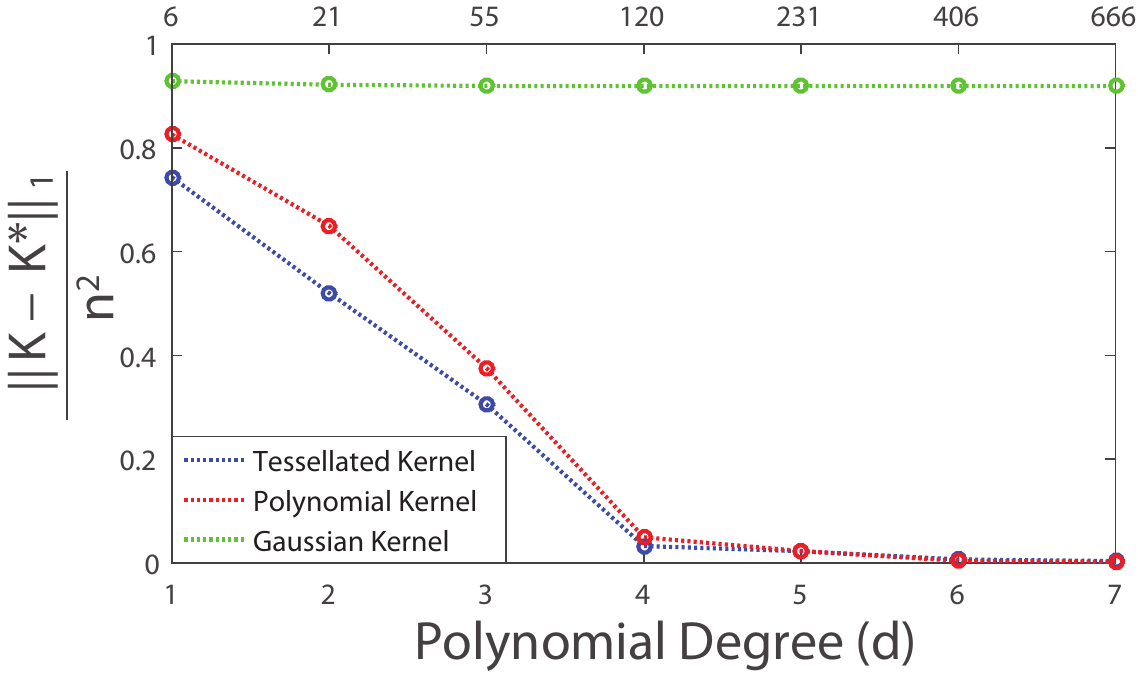}
        \caption{$\frac{\norm{K-K^*}_1}{n^2}$ for the TK and GPK classes of degree $d$ and for a positive combination of $m$ Gaussian kernels.}
    \end{subfigure}%
    ~
    \begin{subfigure}[t]{0.5\textwidth}
        \centering
\includegraphics[width=\textwidth]{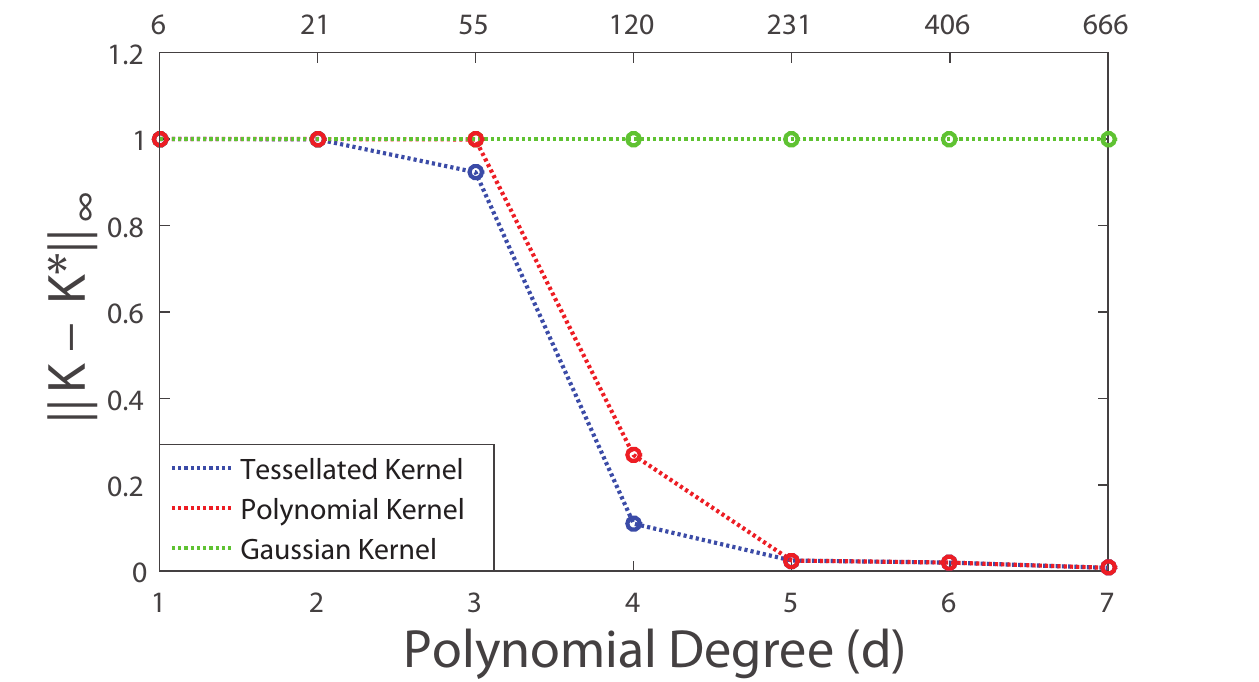}
        \caption{$\norm{K-K^*}_\infty$ for the TK and GPK classes of degree $d$ and for a positive combination of $m$ Gaussian kernels.}
    \end{subfigure}
    \caption{The objective of Optimization Problem~\ref{opt:1norm} and~\ref{opt:infnorm} for the TK and GPK classes of degree $d$ and for a positive combination of $m$ Gaussian kernels with bandwidths ranging from $.01$ to $10$.  The number of bandwidths is selected so that the number of decision variables match in the Gaussian and in the TK kernel case.}\label{fig:kernel_error} \vspace{-5mm}
\end{figure}

We now solve Optimization Problems \eqref{opt:1norm} and \eqref{opt:infnorm} for $\mcl K^\gamma_G$, $\mcl K_P^d$, and $\mcl K^d_T$ as a function of the degree of the polynomials, $d$ and the number of bandwidths selected ($K$). For this test, we use the spiral data set with 20 samples and corresponding labels such that $\sum_{i=1}^m y_i = 0$.
Since half of the entries in $K^*$ are $-1$, and since the Gaussian kernel is globally positive, it is easy to see that for $\mcl K= \mcl K^\gamma_G$ the minimum objective values of Optimization Problems~\eqref{opt:1norm} and \eqref{opt:infnorm} are lower bounded by $0.5$ and $1$ respectively, irrespective of the choice of bandwidths, $\gamma_i$ and number of data points. In Figs.~\ref{fig:kernel_error}(a) and \ref{fig:kernel_error}(b) we numerically show the change in the objective value of Optimization Problems \eqref{opt:1norm} and \eqref{opt:infnorm} for the optimal Gaussian, GPK, and TK kernels as we increase the complexity of the kernel function.  For the TK and GPK kernel functions, we increase the complexity of the kernel function by increasing the degree of the monomial basis while scaling the $x$-axis to ensure equivalent computational complexity.

The results demonstrate that, as expected, the Gaussian kernel saturates with an objective value significantly larger than the lower bound of $0.5$ for the 1-norm and exactly at $1$ for the $\infty$-norm (the projected lower bound). Meanwhile, as the degree increases, both the GPK and TK kernels are able to approximate the kernel matrix arbitrarily well, with almost no error at degree $d=7$. Furthermore, the TK kernels converge somewhat faster.

Note that the optimization problem considered in this subsubsection only concerns the approximation of a kernel function. In Section~\ref{sec:8} we will show that TK kernels also outperform Gaussians when solving the kernel learning SVM Problem~\eqref{OptimalKernelSVM1} for standard data sets.

\section{SDP Formulation of the TK Kernel Learning Algorithm} \label{sec:SDP}\vspace{-1.5mm}
Section \ref{sec:2} gave a convex formulation of the kernel learning problem using the convex constraint $k \in \mcl K$.  Having now defined the TK class of kernels, we now address specific implementations of the TK kernel learning problem using both an SDP method based on Optimization Problem \eqref{SDP} and a method based on the SimpleMKL toolbox. In both cases, our goal for this section is to define an explicit linear map from the elements of the positive matrix variable, $P$, to the values of the kernel function $k(x_i,x_j)$.

To construct our mapping, we first create an index of the elements in the basis $Z_d(z,x)$ which is used in $N_T^d(z,x)$ as defined in Eqn.~\eqref{eqn:N}. Recall $Z_d(z,x)$ is a vector of all monomials of degree $d$ or less of length $q:=\binom{d+2n}{d}$. We now specify that the elements of $Z_d$ are ordered, and by default we use lexicographical ordering on the exponents of the variables of the monomials. Specifically, we denote the jth monomial in $Z_d(z,x)$ where $z,x \in \R^n$ as $z^{\gamma_j}x^{\delta_j} :=\prod_{i=1}^n z_i^{\gamma_{j,i}} x_i^{\delta_{j,i}}$ where $\gamma_j, \delta_j \in \N^n$ and $\{ [\gamma_j,\delta_j] \}_{j=1}^q$ is ordered lexicographically. Note that  $\{ [\gamma_j,\delta_j] \}_{j=1}^q=\{x \in \N^{2n} : \norm{x}_1 \le d\}$. Using this notation, we have the following representation of the TK kernel $k$.

\begin{corr}\label{corr}
Suppose that for $a<b \in \R^n$, $Y={\mcl X}=[a,b]$, and $d \in\N$ we define the finite set $D_d:=\{(\gamma,\delta)\in \N^{2n} : \norm{(\gamma,\delta)}_1 \le d\}$.  Let $\{ [\gamma_i, \delta_i] \}_{i=1}^q \subseteq D_d$ be some ordering of $D_d$ and define $Z_d(z,x)_j = z^{\gamma_j} x^{\delta_j}$. Now let $k$ be as defined in Eqn.~\eqref{eqn:kernel} for some $P>0$ and where $N$ is as defined in Eqn.~\eqref{eqn:N}. If we partition $P = \bmat{Q & R\\ R^T & S}$ then we have,
\[
k(x,y)=\sum_{i,j=1}^q Q_{i,j} g_{i,j}(x,y) + R_{i,j}t_{i,j}(x,y) + R^T_{i,j}t_{i,j}(y,x)  + S_{i,j}  h_{i,j}(x,y)
\]
where $g_{i,j},t_{i,j},h_{i,j}:\R^{n} \times \R^n \rightarrow \R$ are defined as
\begin{align}\label{g}
g_{i,j}(x,y) &:= x^{\delta_i}y^{\delta_j} T(p^*(x,y),b,\gamma_i + \gamma_j + \mathbf{1} ) \\
t_{i,j}(x,y) &:= x^{\delta_{i}}y^{\delta_{j}} T(x,b,\gamma_i + \gamma_j + \mathbf{1}  ) - g_{i,j}(x,y) \notag \\
 h_{i,j}(x,y) &:= x^{\delta_{i}}y^{\delta_{j}} T(a,b,\gamma_i + \gamma_j + \mathbf{1}  ) - g_{i,j}(x,y) - t_{i,j}(x,y) - t_{i,j}(y,x), \notag
\end{align}
where $\mathbf{1} \in \N^n$ is the vector of ones, $p^*:\R^{n}\times \R^n \rightarrow \R^n$ is defined elementwise as
$p^*(x,y)_i = \max \{x_i,y_i \}$, and $T:\R^n \times \R^n \times \N^n\rightarrow \R$ is defined as
\[ T(x,y,\zeta) = \prod_{j=1}^n \left(\hspace{2mm} \frac{y_j^{\zeta_j}}{\zeta_j}-\frac{x_j^{\zeta_j}}{\zeta_j}\right). \]
\end{corr}
\begin{proof}
The proof follows from Theorem~\ref{thm:continuity}.
\end{proof}

An illustration of this map using lexicographical indexing is given in the appendix.  Using a linear map from the elements of $P$ to the value of $k(x,y)$, we may now write the SDP version of the TK kernel learning problem as follows.
\begin{align}\label{SDPOPT}
&\underset{ t \in \R,~\gamma \in \R,~\nu \in \R^{m},~\delta \in \R^m,~Q,R,S \in \R^{q \times q}}{\text{min}} \quad t \\ \nonumber
& \text{subject to:}   \begin{pmatrix}
  G(P) &  e + \nu -  \delta + \gamma y \\ \nonumber
  (e + \nu - \delta + \gamma  y)^T & t-\frac{2}{m \lambda }  \delta^T  e
 \end{pmatrix} \geq 0, \\ \nonumber
 &  \nu \geq 0, \qquad  \delta \geq 0, \qquad P = \bmat{Q & R\\ R^T & S} > 0, \qquad \text{trace}(P) \leq 1, \\ \nonumber
&G_{k,l}(P)= y_ky_l\sum_{i,j=1}^q Q_{i,j} g_{i,j}(x_k,x_l) + R_{i,j}t_{i,j}(x_k,x_l) + R^T_{i,j}t_{i,j}(x_l,x_k)  + S_{i,j}  h_{i,j}(x_k,x_l)
\end{align}

Optimization Problem \eqref{SDPOPT}, then, is an SDP and can, therefore, be solved efficiently using standard SDP solvers such as MOSEK in~\cite{mosek}.   Note that we use the trace constraint to ensure the kernel function is bounded.

Typically SDP problems require roughly $p^2n^2$ number of operations, where $p$ is the number of decision variables and $n$ is the dimension of the SDP constraint (See~\cite{doherty2004complete}). The number of decision variables in~\eqref{SDPOPT} is moderate, increasingly linearly in the number of training data points and the number of elements of $P$. However, this optimization problem has a semi-definite matrix constraint whose dimension is linear in $m$, the number of training data. As we will see in Section~\ref{sec:7}, the increase in training data increases $n$ and limits the amount of training data that can be processed using Optimization Problem~\eqref{SDPOPT}. To improve the scalability of the algorithm, we consider a variation on SimpleMKL.


\section{SimpleMKL Formulation of the TK Kernel Learning Algorithm} \label{sec:6}\vspace{-1.5mm}
Recall that SimpleMKL searches for an optimal positive linear combination of kernel functions from the set~\eqref{simpleMKLSet}.  The algorithm returns a vector of positive weights $\mu$, corresponding to each kernel in the set of a priori selected kernel functions $k_s(x,y)$.  Here we discuss how SimpleMKL as implemented in~\cite{rakotomamonjy_2008} can be used to find optimal combinations of TK kernels.

To create a basis set of TK kernels, we randomly generate a set of $L$ positive semi-definite matrices, $P^s$ for $s=1,\dots,L$ and use SimpleMKL to find the optimal linear combination of the TK kernels defined by each matrix $P^s = \bmat{Q^s & R^s\\ (R^s)^T & S^s}$. Using the basis kernels
\[
k_s(x,y)=\sum_{i,j}^q Q^s_{i,j} g_{i,j}(x,y) + R^s_{i,j}t_{i,j}(x,y)    + R^{s}_{j,i}t_{i,j}(y,x)  + S^s_{i,j}  h_{i,j}(x,y)
\]
where $g(x,y),~t(x,y),~\text{and}~h(x,y)$ are as defined in Eq.~\eqref{g}, we now have
\begin{align} \label{OptimalKernelSVM}
\min_{\mu \ge 0} \max_{\alpha \in \R^m} & \quad \sum_{i=1}^m \alpha_i - \frac{1}{2} \sum_{i,j=1}^m \sum_{s=1}^L \sum_{l,m=1}^{2q} \mu_s \alpha_i \alpha_j y_i y_j k_s(x_i,x_j)  \\ \nonumber
\text{s.t.} & \quad \sum_{i=1}^m \alpha_iy_i = 0, \quad 0 \leq \alpha_i \leq C ~~ \forall ~~ i = 1,...,m.
\end{align}

While the current use of randomly generated matrices is somewhat heuristic, it may be avoided through the development of a dedicated two-step algorithm - wherein the first step optimizes $\alpha$ for a fixed $P$ and the second step fixes $\alpha$ and searches over the positive matrices.

In Section~\ref{sec:7} we will perform a numerical analysis of the complexity of both the SDP and SimpleMKL implementations.

\begin{figure}[t]
    \centering
    \begin{subfigure}[t]{0.45\textwidth}
        \centering
\includegraphics[width=\textwidth]{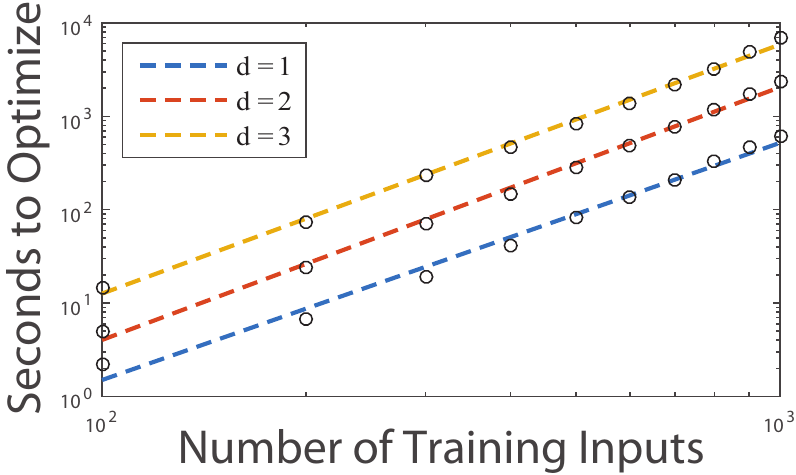}
        \caption{Complexity Scaling for Identification of Circle}
    \end{subfigure}%
    ~
    \begin{subfigure}[t]{0.45\textwidth}
        \centering
\includegraphics[width=\textwidth]{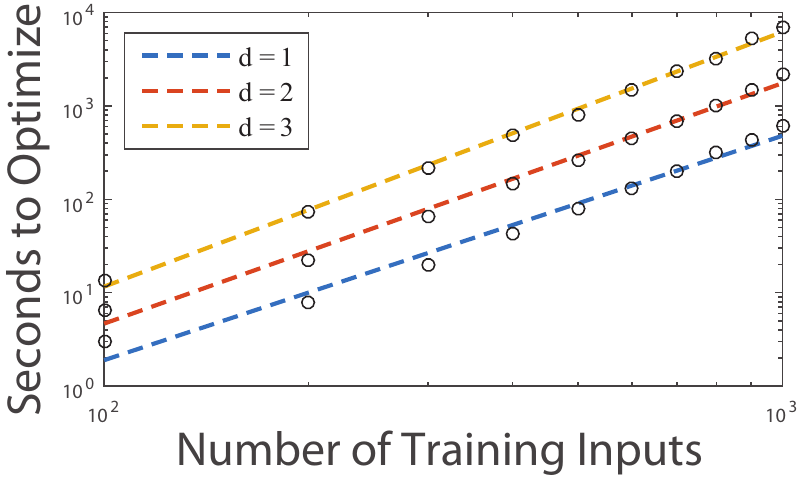}
        \caption{Complexity Scaling for Identification of Spiral}
    \end{subfigure}
    \caption{Log-Log Plot of Computation Time vs number of training data for 2-feature kernel learning.}\label{fig:circle_complexity} \vspace{-5mm}
\end{figure}

\section{Implementation and Complexity Analysis}\label{sec:7}\vspace{-1.5mm}
In this section we first analyze the complexity of Optimization Problem \eqref{SDPOPT} with respect to the number of training points as well as the selected degree of the TK - ${\mcl K}_T^d$.  We then perform the same analysis on Optimization Problem \eqref{OptimalKernelSVM} with respect to the number of training points and the number of random matrices selected.

{\bf Analysis of the SDP Approach:}
In Optimization Problem~\eqref{SDPOPT} the constraint that the kernel be a positive TK kernel can be expressed as an LMI constraint with variables $P_{ij}$. Using Optimization Problem \eqref{SDPOPT}, if $P\in \R^{q \times q}$, and $m$ is the number of training data, with a Mosek implementation, we find experimentally that the complexity of the resulting SDP scales as approximately $m^{2.6}+q^{1.9}$ as can be seen in Fig. \ref{fig:circle_complexity} and is similar to the complexity of other methods such as the hyperkernel approach in~\cite{ong2005learning}. These scaling results are for training data randomly generated by two standard 2-feature example problems (circle and spiral - See Fig.~\ref{fig:spiral_plot}) for degrees $d=1$, $2$, $3$ and where $d$ defines the length of $Z_d$ (and hence $q$) which is the vector of all monomials in 2 variables of degree $d$ or less.

Note that the length of $Z_d$ scales with the degree and number of features, $n$, as $q=\frac{(n+d-1)!}{n!d!}$.  For a large number of features and a high degree, the size of $Z_d$ will become unmanageably large. Note, however, that, as indicated in Section~\ref{sec:4}, even when $d=0$, every TK kernel is universal.

\begin{figure}[t]
    \centering
    \begin{subfigure}[t]{0.45\textwidth}
        \centering
\includegraphics[trim=15 0 65 0,clip,width=\textwidth]{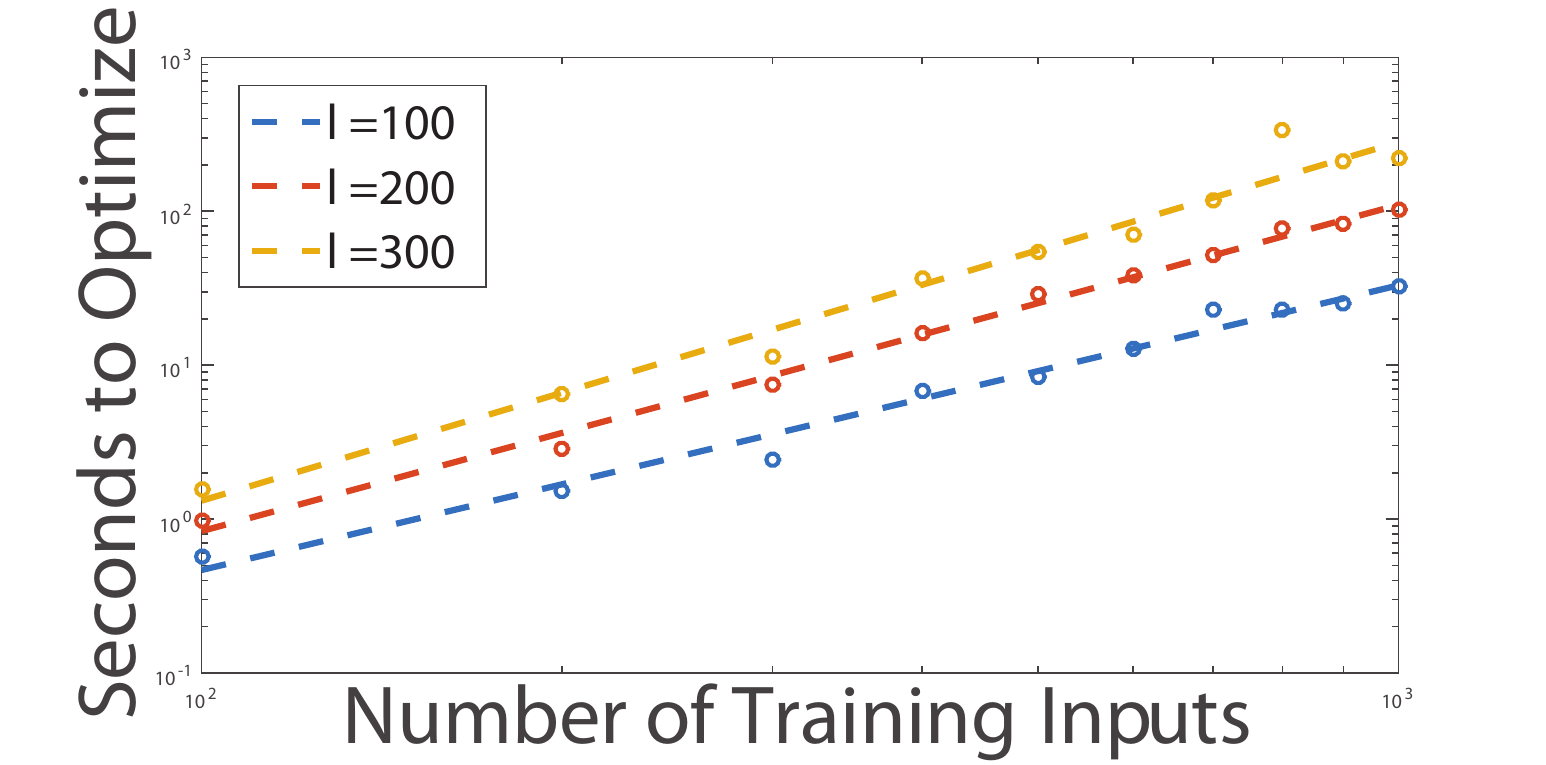}
        \caption{Complexity Scaling for Identification of Circle using SimpleMKL with TK kernels.}
    \end{subfigure}%
    ~
    \begin{subfigure}[t]{0.45\textwidth}
        \centering
\includegraphics[trim=15 0 65 0,clip,width=\textwidth]{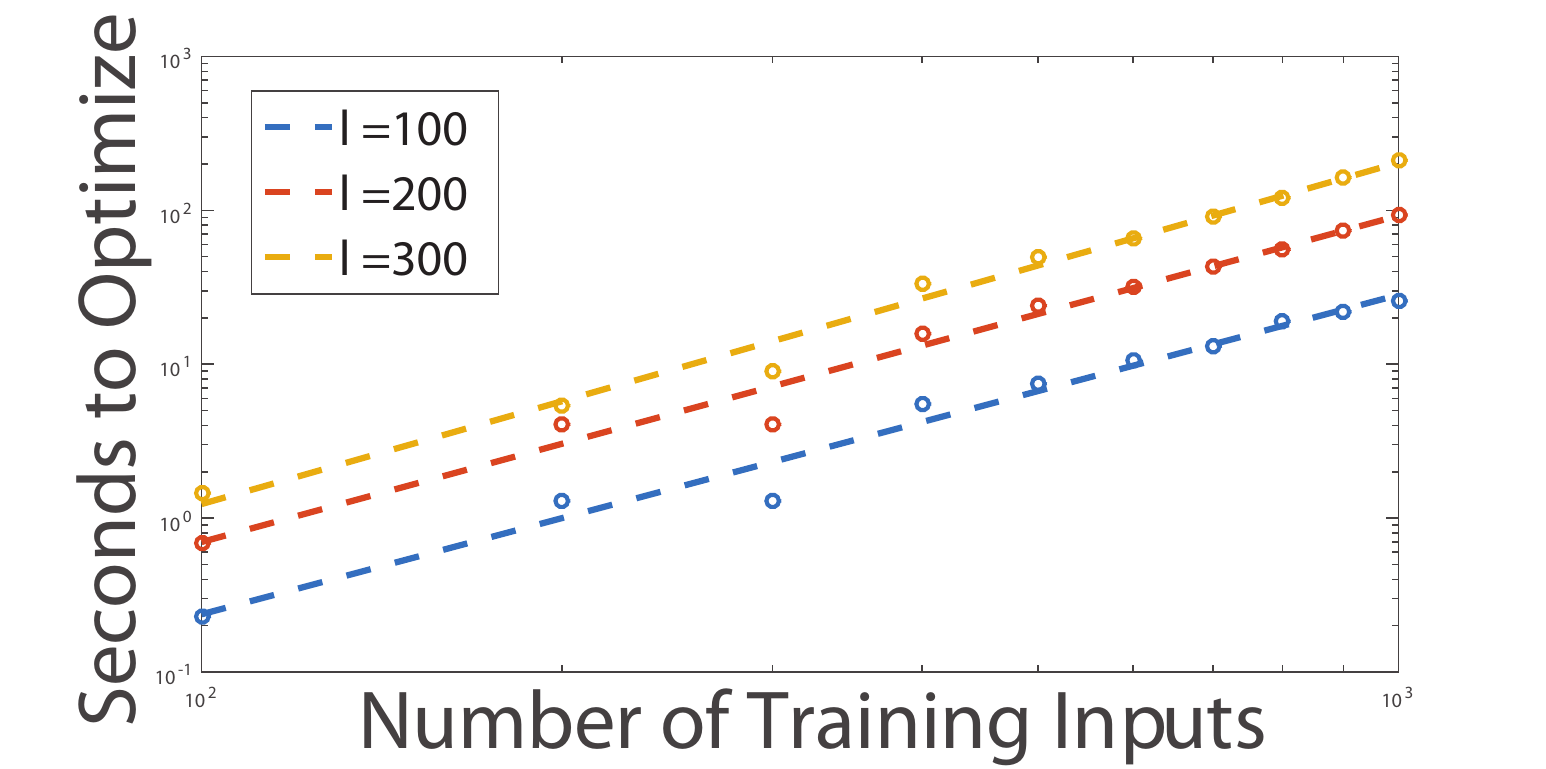}
        \caption{Complexity Scaling for Identification of Spiral using SimpleMKL with TK kernels.}
    \end{subfigure}
    \caption{Log-Log Plot of Computation Time vs number of training data for 2-feature kernel learning using SimpleMKL with TK kernels.}\label{fig:SMKL_complexity} \vspace{-5mm}
\end{figure}

{\bf Analysis of the SimpleMKL Approach:}
Solving Optimization Problem~\eqref{OptimalKernelSVM} with SimpleMKL we first need to generate a set of random matrices.  If we have $L$ random positive semi-definite matrices and $m$ training data points then we find experimentally that the complexity of the resulting SDP scales as approximately $m^{2.1}+L^{1.6}$ as can be seen in Fig. \ref{fig:SMKL_complexity}. These scaling results are, as in the results for the SDP method, for training data randomly generated by two standard 2-feature example problems (circle and spiral - See Fig.~\ref{fig:spiral_plot}).  We select the number of training data $m$, to vary between 100 and 1000 points and select the number of random matrices to be $L = 100,200,300$.

Note that the complexity of the SimpleMKL version is largely independent of the selected degree of the polynomial.  However, a larger degree means that the matrices $P$ are larger, and therefore a larger number of random positive semi-definite matrices, $L$, should be selected.
~\\
~\\
~\\
\section{Accuracy and Comparison With Existing Methods}\label{sec:8}\vspace{-2mm}
\begin{wrapfigure}{h}{0.43\textwidth}
\vspace{-5mm}
\centering
\includegraphics[width=.45\textwidth]{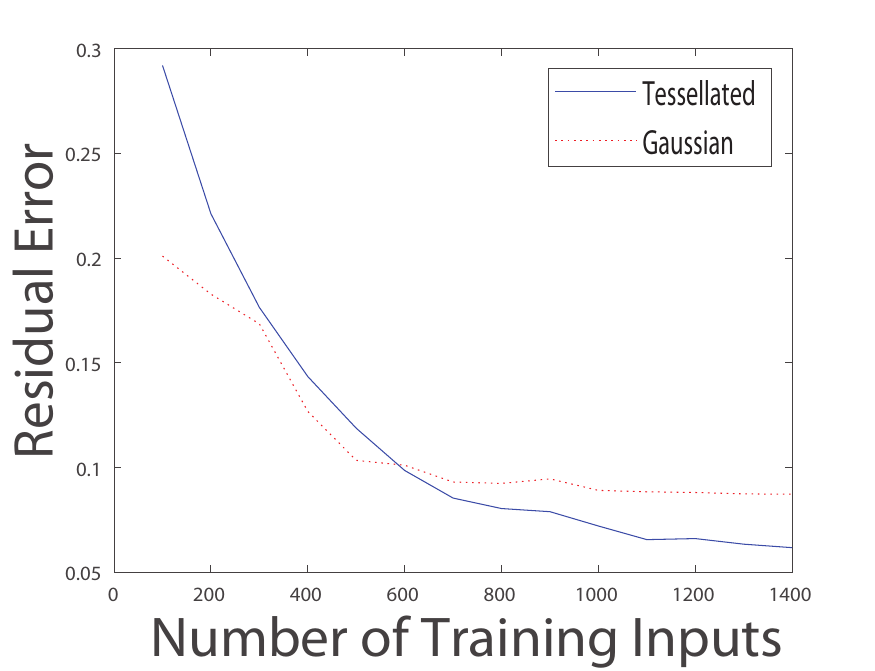}
\caption{TSA compared with SimpleMKL for spiral data set with artificial additive noise.}\vspace{-5mm}\label{fig:error_spiral_noise}
\end{wrapfigure}
 In this section, we evaluate the relative accuracy of Optimization Problem~\eqref{SDPOPT} using SDP and Optimization Problem~\eqref{OptimalKernelSVM} using SimpleMKL.
To evaluate the accuracy, we applied 5 variations of the kernel learning problem to 5 randomly selected benchmark data sets from the UCI Machine learning Data Repository - Liver, Cancer, Heart, Pima, and Ionosphere.
In all evaluations of Test Set Accuracy (TSA), the data is partitioned into 80\% training data and 20\% testing and this partition is repeated 30 times to obtain 30 sets of training and testing data. For all numerical tests we use the soft-margin problem with regularization parameter $C$, where $C$ is selected from a set of values picked a priori by 5-fold cross-validation.  To perform 5-fold cross-validation we split the training data set into five groups, solve the optimization problem using each potential value of $C$ on four of the five groups and test the optimal classifier performance on the remaining group.  We repeat this process using each of the five groups as the test set and select the value of $C$ which led to the best average performance.

The 5 variations on the kernel learning problem are

\noindent\textbf{[TK]} We use the SDP algorithm in~\eqref{SDPOPT} using $d=1$ (Except Ionosphere, which uses $d=0$) but set $\gamma = 0$ to decrease numerical complexity. To determine the integral in~\eqref{SDPOPT}, we first scaled the data so that $x_i \in [0,1]^n$, and then set ${\mcl X} := [0-\epsilon,1+\epsilon]^n$, where $\epsilon>0$ was chosen by 5-fold cross-validation.

\noindent\textbf{[SimpleMKL]} We use SimpleMKL with a standard selection of Gaussian and polynomial kernels with bandwidths arbitrarily chosen between .5 and 10 and polynomial degrees one through three - yielding approximately $13(n+1)$ kernels;

\noindent\textbf{[SimpleMKL TK]} We randomly generated a sequence of $300$ positive semidefinite matrices and use these as the SimpleMKL library of kernels;

\noindent\textbf{[SimpleMKL TK+]} We combined the libraries in [SimpleMKL] and [SimpleMKL TK] into a single SimpleMKL implementation;

\noindent\textbf{[Neural Net]} We use 3 layer neural network with 50 hidden layers using MATLABs (\texttt{patternnet}) implementation.

In Table~\ref{sample-table}, we see the average TSA for these four approaches as applied to several randomly selected benchmark data sets from the UCI Machine learning Data Repository. In all cases, either [TK] or [SimpleMKL TK] met or in some cases significantly exceeded the accuracy of [SimpleMKL].

\begin{figure}[t]
    \centering
    \begin{subfigure}[t]{0.23\textwidth}
        \centering
\includegraphics[width=\textwidth]{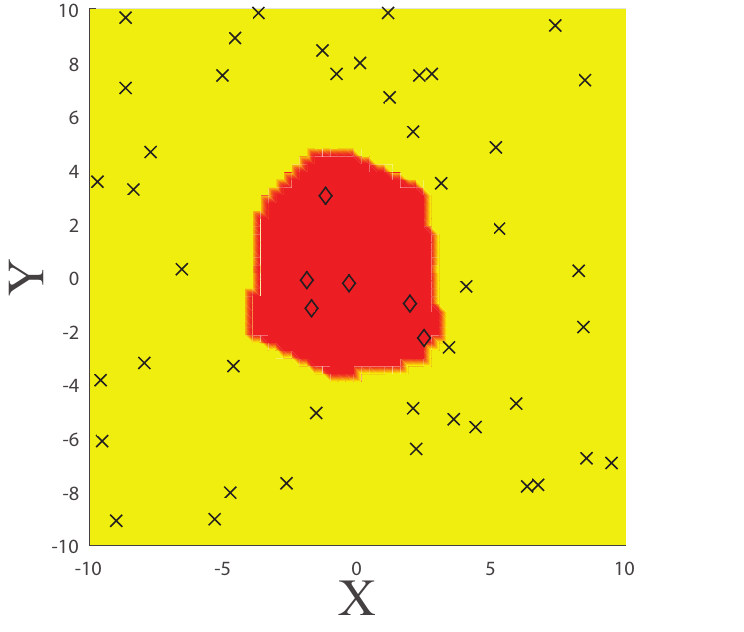}
        \caption{Circle data set classified with [TK] (n=50)}
    \end{subfigure}%
    ~
    \begin{subfigure}[t]{0.23\textwidth}
        \centering
\includegraphics[width=\textwidth]{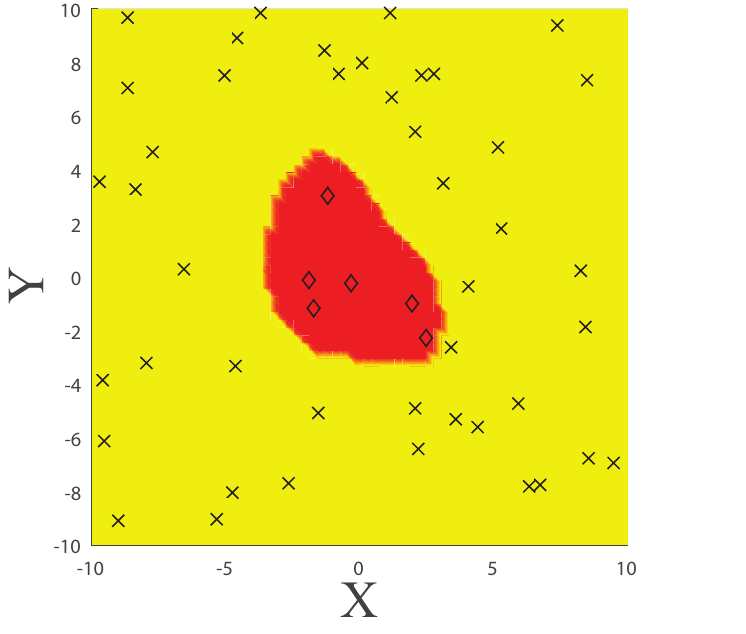}
        \caption{Circle data set classified with [SimpleMKL] (n=50)}
    \end{subfigure}
    ~
    \begin{subfigure}[t]{0.23\textwidth}
        \centering
\includegraphics[width=\textwidth]{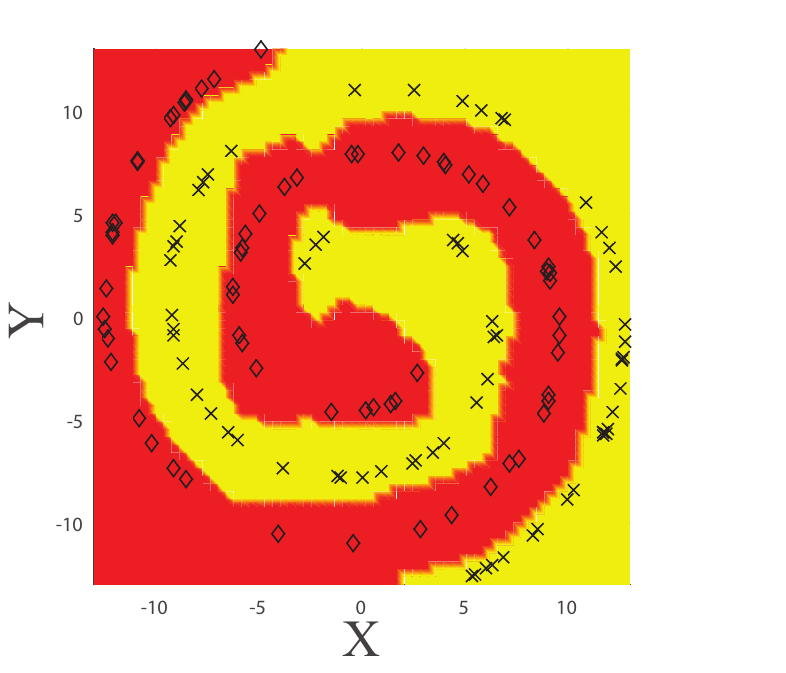}
        \caption{Spiral data set classified with [TK] (n=150)}
    \end{subfigure}%
    ~
    \begin{subfigure}[t]{0.23\textwidth}
        \centering
\includegraphics[width=.82\textwidth]{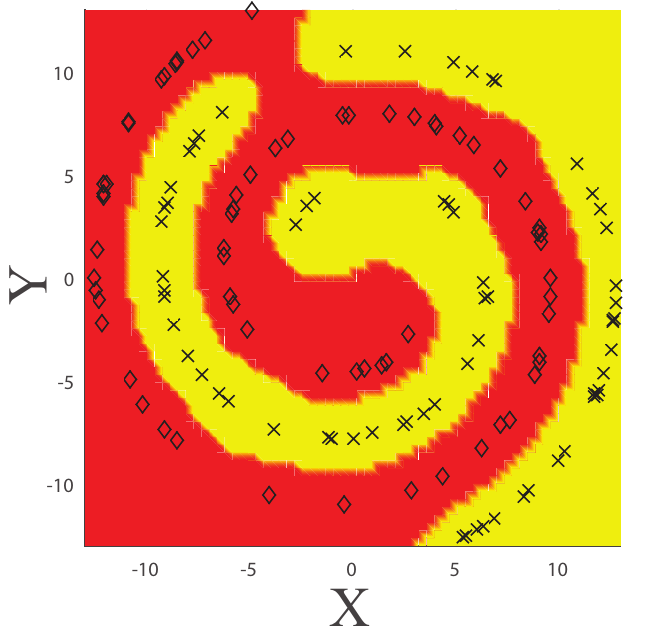}
        \caption{Circle data set classified with [SimpleMKL] (n=150)}
    \end{subfigure}
    \caption{Discriminant Surface for Circle and Spiral Separator using method [TK] as Compared with [SimpleMKL] for $n$ training data.}\label{fig:spiral_plot}
\end{figure}

\begin{table*}[t]
  \caption{TSA comparison for algorithms [TK], [SimpleMKL], [SimpleMKL TK], [SimpleMKL TK+], and [Neural Net].  The maximum TSA for each data set is bold. The average TSA, standard deviation of TSA and time to compute are shown below.  $m$ is size of data set and $n$ the number of features.}
  \label{sample-table}
  \centering
\begin{tabular}{  c |  c  c c c } \hline
Data Set & Method        &    Accuracy                & Time   &  Data Features\\ \hline
       &  TK    &    {\bf 72.32 $\pm$ 4.92}  & 95.75 $\pm$ 2.68 & \\
Liver  &  SimpleMKL      &    65.51 $\pm$ 5.10        & 2.61  $\pm$ 0.42 & m = 346\\
       &  SimpleMKL TK &    70.58 $\pm$ 4.69        & 8.37 $\pm$ 0.30 & n = 6\\
       &  SimpleMKL TK+       &    70.53 $\pm$ 4.79        & 14.70 $\pm$ 0.76 & \\
       &  Neural Net     &    66.32 $\pm$ 7.46        & 0.14 $\pm$ 0.04 & \\ \hline
       &  TK    &    {\bf 97.18 $\pm$ 1.48}  & 636.17 $\pm$ 25.43  & \\
Cancer &  SimpleMKL      &    96.55 $\pm$ 1.34        & 14.74 $\pm$ 1.33 & m = 684 \\
       &  SimpleMKL TK &    96.89 $\pm$ 1.43        & 45.84 $\pm$ 4.28 & n = 9\\
       &  SimpleMKL TK+       &    96.89 $\pm$ 1.42        & 65.08 $\pm$ 10.52 & \\
       &  Neural Net     &    96.67 $\pm$ 1.30        & 0.18 $\pm$ 0.06 & \\ \hline
       &  TK    &    83.46 $\pm$ 4.56        & 221.67 $\pm$ 29.63 & \\
Heart  &  SimpleMKL      &    83.70 $\pm$ 4.77        & 3.09 $\pm$ 0.19 & m = 271\\
       &  SimpleMKL TK&    {\bf 84.38 $\pm$ 4.34 } & 55.48 $\pm$ 2.67 & n = 13 \\
       &  SimpleMKL TK+       &    83.64 $\pm$ 4.54        & 13.23 $\pm$ 2.70 & \\
       &  Neural Net     &    78.64 $\pm$ 5.19        & 0.12 $\pm$ 0.01 & \\ \hline
       &  TK    &    76.32 $\pm$ 3.10        & 1211.66 $\pm$ 27.01 & \\
Pima   &  SimpleMKL      &    76.00 $\pm$ 3.33        & 19.04  $\pm$ 2.33 & m=769\\
       &  SimpleMKL TK &    {\bf 76.75 $\pm$ 2.81}  & 34.65 $\pm$ 23.28 & n = 8\\
       &  SimpleMKL TK+       &    76.57 $\pm$ 2.72        & 96.20 $\pm$ 30.42 & \\
       &  Neural Net     &    75.35 $\pm$ 2.98        & 0.24 $\pm$ 0.19 & \\
       \hline
       &  TK    &    {\bf 93.24 $\pm$ 3.04}  & 6.69 $\pm$ 0.27 & \\
Ionosphere &  SimpleMKL  &    92.16 $\pm$ 2.78        & 26.24 $\pm$ 2.78 & m = 352\\
       &  SimpleMKL TK &    87.65 $\pm$ 2.88        & 8.28 $\pm$ .16 &   n = 34\\
       &  SimpleMKL TK+       &    92.16 $\pm$ 2.78        & 50.77 $\pm$ 2.98 & \\
       &  Neural Net     &    90.85 $\pm$ 3.42        & 0.16 $\pm$ 0.02 & \\ \hline

\end{tabular}
\end{table*}

\begin{figure}[t]
    \centering
    \begin{subfigure}[t]{0.45\textwidth}
        \centering
\includegraphics[width=\textwidth]{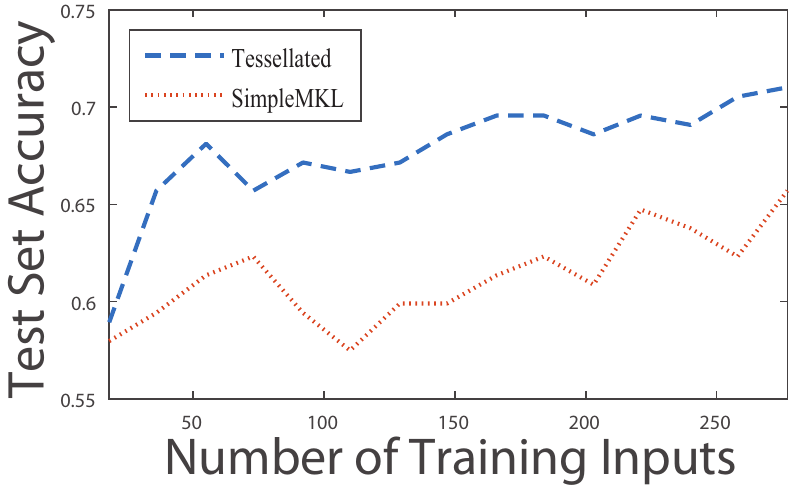}
        \caption{Average test set accuracy on the Liver data set vs. the number of training data for [TK] compared to [SimpleMKL] }
    \end{subfigure}\label{fig:TSA_liver}
    ~
    \begin{subfigure}[t]{0.45\textwidth}
        \centering
\includegraphics[width=\textwidth]{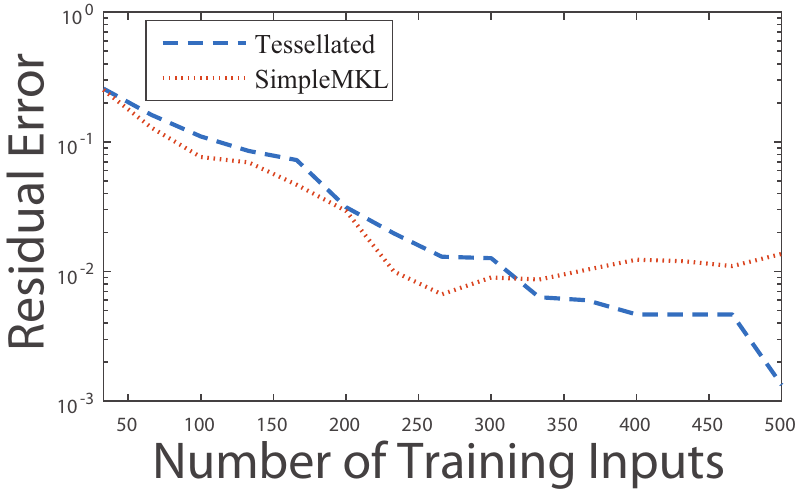}
        \caption{Semilog plot of residual error on generated 2D spiral data vs. number of training data for [TK] compared to [SimpleMKL]. }
    \end{subfigure}\label{fig:error_spiral}
    \caption{Plots demonstrating the change in accuracy of [TK] and [SimpleMKL] with respect to the number of training inputs.  The residual error is defined as 1-TSA where TSA is the test set accuracy.}
\end{figure}\label{fig:error}

In addition to the standard battery of tests, we performed a secondary analysis to demonstrate the advantages of the TK class when the ratio of training data to the number of features is high. For this analysis, we use the liver data set (6 features) and the spiral discriminant (2 features) from \cite{lang_1988} (we also briefly examine the unit circle). For the liver data set, in Figure~\ref{fig:error}, we see a semilog plot of the residual error (i.e. 1-TSA) as the size of the training data increases as compared with SimpleMKL. This figure shows consistent improvement of [TK] over standard usage of [SimpleMKL]. For the spiral case, in Figure~\ref{fig:error} we again see a semilog plot of the residual error as the size of the training data increases as compared with [SimpleMKL]. In this case, both methods converge well with [TK] showing significant improvement over [SimpleMKL] only for very large training data sets.

To explore how increasing the number of training data affects the classifier, we generated a new 1400 point training data set with additive noise of zero mean and $\sigma = .1$. The results are seen in Figure~\ref{fig:error_spiral_noise}. In this case, we see that [TK] significantly outperform [SimpleMKL] beginning at 600 data points.

Finally, as an illustration, we plotted the discriminant surface for both the spiral and unit circle data sets using both the [TK] and [SimpleMKL] methods using 150 training data points. These 2D surfaces are found in Figure~\ref{fig:spiral_plot}. \vspace{-2mm}

\section{Conclusion}\vspace{-3mm}
In this paper, we have proposed a new class of universal kernel functions. This set of kernels can be parameterized directly using positive matrices or indirectly using positive coefficients combined with randomly generated positive matrices. Furthermore, any element of this class is universal in the sense that the hypothesis space is dense in $L_2$, giving it comparable performance to and properties of the Gaussian kernels. However, unlike the Gaussian or RBFs, the TK class does not require a set of bandwidths to be chosen a priori. Furthermore, by increasing the degree of the monomial basis, we have shown that the TK class can approximate any kernel matrix arbitrarily well.

We have demonstrated the effectiveness of the TK class on several data sets from the UCI repository.  We have shown that the computational complexity is comparable to other SDP-based kernel learning methods. Furthermore, by using a randomized basis for the positive matrices, we have shown that the TK class can be readily integrated with existing multiple kernel learning algorithms such as SimpleMKL - yielding similar results with less computational complexity. In most cases, either the optimal TK kernel or the MKL learned sub-optimal TK kernel will outperform or match an MKL approach using Gaussian and polynomial kernels with respect to the Test Set Accuracy.  Finally, we note that this universal class of kernels can be trivially extended to matrix-valued kernels for use in, e.g. multi-task learning as in~\cite{caponnetto_2008}.

\acks
This work was supported by NSF Grant 026257-001.

\newpage
\appendix
\section{Example Calculation of g, h, and t}
In this appendix, we illustrate the lexicographical ordering $\{[\gamma_i,\delta_i]\}$ and the maps $g_{i,j}(x,y),$ $t_{i,j}(x,y),$ $t_{i,j}(y,x)$ and $h_{i,j}(x,y)$ for the special case of
$a = \bmat{0\\0}$, $b = \bmat{3\\4}$.  We choose $d = 1$ which implies $Z_d$ is length $q=5$.  This yields
\[Z_1(z,x) =  \left[ \begin{array}{c}
1 \\
x_1  \\
x_2 \\
z_1 \\
z_2 \end{array} \right],\]
The associated $\delta$ and $\gamma$ are then given by
 \[ \delta =  \left[ \begin{array}{cc}
0 & 0 \\
1 & 0 \\
0 & 1  \\
0 & 0 \\
0 & 0  \end{array} \right], \quad
\gamma =  \left[ \begin{array}{cc}
0 & 0 \\
0 & 0 \\
0 & 0 \\
1 & 0 \\
0 & 1 \end{array} \right]. \]

For brevity we will assemble the matrices $g_{i,j}(x,y),~t_{i,j}(x,y),~t_{i,j}(y,x)$ and $h_{i,j}(x,y)$ for the first, second and fourth elements of $Z_1(z,x)$.

Assembling $g_{i,j}(x,y),~t_{i,j}(x,y),~t_{i,j}(y,x)$ and $h_{i,j}(x,y)$ for $i,j=1,2,4$ into matrices of size $\R^{3 \times 3}$, and by using Equation~\eqref{g} in Corollary~\ref{corr}, we have that,
\begin{align*}
g(x,y)=&\left[ \begin{array}{c c c}
(3-p^*_1)(4-p^*_2) & y_1(3-p^*_1)(4-p^*_2)  & \frac{1}{2}(9-(p^*_1)^2)(4-p^*_2) \\
x_1(3-p^*_1)(4-p^*_2)  & x_1 y_1(3-p^*_1)(4-p^*_2)  & x_1\frac{1}{2}(9-(p^*_1)^2)(4-p^*_2)\\
\frac{1}{2}(9-(p^*_1)^2)(4-p^*_2) & y_1\frac{1}{2}(9-(p^*_1)^2)(4-p^*_2) & \frac{1}{3}(27-(p^*_1)^3)(4-p^*_2) \end{array} \right], \\
t(x,y)=&\left[ \begin{array}{c c c}
(3-x_1)(4-x_2) & y_1(3-x_1)(4-x_2)  & \frac{1}{2}(9-x_1^2)(4-x_2) \\
x_1(3-x_1)(4-x_2)  & x_1 y_1(3-x_1)(4-x_2)  & x_1\frac{1}{2}(9-x_1^2)(4-x_2)\\
\frac{1}{2}(9-x_1^2)(4-x_2) & y_1\frac{1}{2}(9-x_1^2)(4-x_2) & \frac{1}{3}(27-x_1^3)(4-x_2) \end{array} \right] - g(x,y), \\
t(y,x)=&\left[ \begin{array}{c c c}
(3-y_1)(4-y_2) & x_1(3-y_1)(4-y_2)  & \frac{1}{2}(9-y_1^2)(4-y_2) \\
y_1(3-y_1)(4-y_2)  & x_1 y_1(3-y_1)(4-y_2)  & y_1\frac{1}{2}(9-y_1^2)(4-y_2)\\
\frac{1}{2}(9-y_1^2)(4-y_2) & x_1\frac{1}{2}(9-y_1^2)(4-y_2) & \frac{1}{3}(27-y_1^3)(4-y_2) \end{array} \right] - g(x,y), \\
h(x,y)=&\left[ \begin{array}{c c c}
12 & 12y_1 & 36 \\
12x_1 & 12 x_1 y_1 & 36x_1 \\
36 & 36y_1 & 108 \end{array} \right]-g(x,y)-t(x,y)-t(y,x),
\end{align*}
where we have defined $p^*$ element wise as $p^*_i = \text{max}\{x_i,y_i\}$.

\vskip 0.2in
\bibliographystyle{plain}

\begin{thebibliography}{32}
\providecommand{\natexlab}[1]{#1}
\providecommand{\url}[1]{\texttt{#1}}
\expandafter\ifx\csname urlstyle\endcsname\relax
  \providecommand{\doi}[1]{doi: #1}\else
  \providecommand{\doi}{doi: \begingroup \urlstyle{rm}\Url}\fi

\bibitem[Alizadeh et~al.(1998)Alizadeh, Haeberly, and Overton]{alizadeh_1998}
F.~Alizadeh, J.-P. Haeberly, and M.~Overton.
\newblock Primal-dual interior-point methods for semidefinite programming:
  convergence rates, stability and numerical results.
\newblock \emph{SIAM Journal on Optimization}, 1998.

\bibitem[ApS(2015)]{mosek}
MOSEK ApS.
\newblock \emph{The MOSEK optimization toolbox for MATLAB manual. Version 7.1
  (Revision 28).}, 2015.

\bibitem[Borgwardt et~al.(2006)Borgwardt, Gretton, Rasch, Kriegel,
  Sch{\"o}lkopf, and Smola]{borgwardt2006}
K.~M. Borgwardt, A.~Gretton, M.~J. Rasch, H.~Kriegel, B.~Sch{\"o}lkopf, and
  A.~J. Smola.
\newblock Integrating structured biological data by kernel maximum mean
  discrepancy.
\newblock \emph{Bioinformatics}, 2006.

\bibitem[Caponnetto et~al.(2008)Caponnetto, Micchelli, Pontil, and
  Ying]{caponnetto_2008}
A.~Caponnetto, C.~Micchelli, M.~Pontil, and Y.~Ying.
\newblock Universal multi-task kernels.
\newblock \emph{Journal of Machine Learning Research}, 2008.

\bibitem[Collins and Duffy(2002)]{collins_2002}
M.~Collins and N.~Duffy.
\newblock Convolution kernels for natural language.
\newblock In \emph{Advances in Neural Information Processing Systems}, 2002.

\bibitem[Cortes et~al.(2009)Cortes, Mohri, and
  Rostamizadeh]{cortes2009learning}
C.~Cortes, M.~Mohri, and A.~Rostamizadeh.
\newblock Learning non-linear combinations of kernels.
\newblock In \emph{Advances in Neural Information Processing Systems}, 2009.

\bibitem[Cortes et~al.(2012)Cortes, Mohri, and Rostamizadeh]{cortes_2012}
C.~Cortes, M.~Mohri, and A.~Rostamizadeh.
\newblock Algorithms for learning kernels based on centered alignment.
\newblock \emph{Journal of Machine Learning Research}, 2012.

\bibitem[Doherty et~al.(2004)Doherty, Parrilo, and
  Spedalieri]{doherty2004complete}
A.~C. Doherty, P.~A. Parrilo, and F.~M. Spedalieri.
\newblock Complete family of separability criteria.
\newblock \emph{Physical Review A}, 2004.

\bibitem[Eskin et~al.(2003)Eskin, Weston, Noble, and Leslie]{eskin_2003}
E.~Eskin, J.~Weston, W.~Noble, and C.~Leslie.
\newblock Mismatch string kernels for {SVM} protein classification.
\newblock In \emph{Advances in Neural Information Processing Systems}, 2003.

\bibitem[Gai et~al.(2010)Gai, Chen, and Zhang]{gai2010learning}
K.~Gai, G.~Chen, and C.~S. Zhang.
\newblock Learning kernels with radiuses of minimum enclosing balls.
\newblock In \emph{Advances in Neural Information Processing Systems}, 2010.

\bibitem[G{\"a}rtner et~al.(2003)G{\"a}rtner, Flach, and Wrobel]{gartner_2003}
T.~G{\"a}rtner, P.~Flach, and S.~Wrobel.
\newblock On graph kernels: Hardness results and efficient alternatives.
\newblock In \emph{Learning Theory and Kernel Machines}. 2003.

\bibitem[Gasca and Sauer(2001)]{gasca_2001}
M.~Gasca and T.~Sauer.
\newblock On the history of multivariate polynomial interpolation.
\newblock In \emph{Numerical Analysis: Historical Developments in the 20th
  Century}. 2001.

\bibitem[G{\"o}nen and Alpaydin(2008)]{gonen_2008}
M.~G{\"o}nen and E.~Alpaydin.
\newblock Localized multiple kernel learning.
\newblock In \emph{Proceedings of the International Conference on Machine
  learning}, 2008.

\bibitem[G{\"o}nen and Alpayd{\i}n(2011)]{gonen2011multiple}
M.~G{\"o}nen and E.~Alpayd{\i}n.
\newblock Multiple kernel learning algorithms.
\newblock \emph{Journal of Machine Learning Research}, 2011.

\bibitem[Haussler(1999)]{haussler_1999}
D.~Haussler.
\newblock Convolution kernels on discrete structures.
\newblock Technical report, University of California in Santa Cruz, 1999.

\bibitem[Jain et~al.(2012)Jain, Vishwanathan, and Varma]{jain_2012}
A.~Jain, S.~Vishwanathan, and M.~Varma.
\newblock {SPF-GMKL}: generalized multiple kernel learning with a million
  kernels.
\newblock In \emph{Proceedings of the ACM International Conference on Knowledge
  Discovery and Data Mining}, 2012.

\bibitem[Lanckriet et~al.(2004)Lanckriet, Cristianini, Bartlett, El~Ghaoui, and
  Jordan]{lanckriet_2004}
G.~Lanckriet, N.~Cristianini, P.~Bartlett, L.~El~Ghaoui, and M.~Jordan.
\newblock Learning the kernel matrix with semidefinite programming.
\newblock \emph{Journal of Machine Learning Research}, 2004.

\bibitem[Lang(1988)]{lang_1988}
K.~Lang.
\newblock Learning to tell two spirals apart.
\newblock In \emph{Proceedings of the Connectionist Models Summer School},
  1988.

\bibitem[Lodhi et~al.(2002)Lodhi, Saunders, Shawe-Taylor, Cristianini, and
  Watkins]{lodhi_2002}
H.~Lodhi, C.~Saunders, J.~Shawe-Taylor, N.~Cristianini, and C.~Watkins.
\newblock Text classification using string kernels.
\newblock \emph{Journal of Machine Learning Research}, 2002.

\bibitem[Micchelli et~al.(2006)Micchelli, Xu, and Zhang]{micchelli_2006}
C.~Micchelli, Y.~Xu, and H.~Zhang.
\newblock Universal kernels.
\newblock \emph{Journal of Machine Learning Research}, 2006.

\bibitem[Ong et~al.(2005)Ong, Smola, and Williamson]{ong2005learning}
C.~S. Ong, A.~J. Smola, and R.~C. Williamson.
\newblock Learning the kernel with hyperkernels.
\newblock \emph{Journal of Machine Learning Research}, 2005.

\bibitem[Peet et~al.(2009)Peet, Papachristodoulou, and Lall]{peet_SICON_2009}
M.~M. Peet, A.~Papachristodoulou, and S.~Lall.
\newblock Positive forms and stability of linear time-delay systems.
\newblock \emph{SIAM Journal on Control and Optimization}, 2009.

\bibitem[Rahimi and Recht(2008)]{rahimi_2008}
A.~Rahimi and B.~Recht.
\newblock Random features for large-scale kernel machines.
\newblock In \emph{Advances in neural information processing systems}, pages
  1177--1184, 2008.

\bibitem[Rakotomamonjy et~al.(2008)Rakotomamonjy, Bach, Canu, and
  Grandvalet]{rakotomamonjy_2008}
A.~Rakotomamonjy, F.~R. Bach, S.~Canu, and Y.~Grandvalet.
\newblock Simple{MKL}.
\newblock \emph{Journal of Machine Learning Research}, 2008.

\bibitem[Recht(2006)]{recht2006convex}
B.~Recht.
\newblock \emph{Convex Modeling with Priors}.
\newblock PhD thesis, Massachusetts Institute of Technology, 2006.

\bibitem[Sch{\"o}lkopf et~al.(2002)Sch{\"o}lkopf, Smola, and
  Bach]{LearningWithKernels}
B.~Sch{\"o}lkopf, A.~J. Smola, and F.~Bach.
\newblock \emph{Learning with kernels: support vector machines, regularization,
  optimization, and beyond}.
\newblock MIT press, 2002.

\bibitem[Shekhtman(1982)]{shekhtman1982piecewise}
B.~Shekhtman.
\newblock Why piecewise linear functions are dense in {C} [0, 1].
\newblock \emph{Journal of Approximation Theory}, 1982.

\bibitem[Sonnenburg et~al.(2010)Sonnenburg, Henschel, Widmer, Behr, Zien, Bona,
  Binder, Gehl, Franc, et~al.]{SHOGUN}
S.~{\'C}. Sonnenburg, S.~Henschel, C.~Widmer, J.~Behr, A.~Zien, F.~de Bona,
  A.~Binder, C.~Gehl, V.~Franc, et~al.
\newblock The {SHOGUN} machine learning toolbox.
\newblock \emph{Journal of Machine Learning Research}, 2010.

\bibitem[Subrahmanya and Shin(2010)]{subrahmanya_2010}
N.~Subrahmanya and Y.~Shin.
\newblock Sparse multiple kernel learning for signal processing applications.
\newblock \emph{IEEE Transactions on Pattern Analysis and Machine
  Intelligence}, 2010.

\bibitem[Sun(2005)]{sun_2005}
H~Sun.
\newblock Mercer theorem for {RKHS} on noncompact sets.
\newblock \emph{Journal of Complexity}, 2005.

\bibitem[Wang et~al.(2013)Wang, Xiao, and Zhou]{wang2013}
H.~Wang, Q.~Xiao, and D.~Zhou.
\newblock An approximation theory approach to learning with $\ell_1$
  regularization.
\newblock \emph{Journal of Approximation Theory}, 2013.

\bibitem[Zanaty and Afifi(2011)]{zanaty_2011}
E.~Zanaty and A.~Afifi.
\newblock Support vector machines {(SVMs)} with universal kernels.
\newblock \emph{Applied Artificial Intelligence}, 2011.

\end{thebibliography}

\end{document}